\documentclass[conference]{IEEEtran}
\ifCLASSINFOpdf
\else
\fi
\usepackage{amsmath}
\usepackage[compatibility=false]{caption}
\usepackage{subcaption}
\usepackage{graphicx}
\usepackage[ruled,vlined]{algorithm2e}
\usepackage{booktabs} 
\usepackage[colorlinks=true, linkcolor=blue, citecolor=blue, urlcolor=blue]{hyperref}
\usepackage{fancyhdr}
\begin{document}
%
\title{PQFed: A \underline{P}rivacy-Preserving \underline{Q}uality-Controlled \underline{Fed}erated Learning Framework}


\author{

\IEEEauthorblockN{Weiqi Yue}
\IEEEauthorblockA{Case Western Reserve University\\ wxy215@case.edu}
\and
\IEEEauthorblockN{Wenbiao Li}
\IEEEauthorblockA{Case Western Reserve University\\ wxl387@case.edu}
\and
\IEEEauthorblockN{Yuzhou Jiang}
\IEEEauthorblockA{Case Western Reserve University\\ yxj466@case.edu}
\and
\IEEEauthorblockN{Anisa Halimi}
\IEEEauthorblockA{IBM Research Europe\\ Anisa.Halimi@ibm.com}
\and
\IEEEauthorblockN{Roger H. French}
\IEEEauthorblockA{Case Western Reserve University\\ rxf131@case.edu}
\and
\IEEEauthorblockN{Erman Ayday}
\IEEEauthorblockA{Case Western Reserve University\\ exa208@case.edu}
}

\maketitle

\begin{abstract}
Federated learning enables collaborative model training without sharing raw data, but data heterogeneity consistently challenges the performance of the global model. 
Traditional optimization methods often rely on collaborative global model training involving all clients, followed by local adaptation to improve individual performance. 
In this work, we focus on early-stage quality control and propose PQFed, a novel and privacy-preserving personalized federated learning framework that designs customized training strategies for each client prior to the federated training process.
PQFed extracts representative features from each client’s raw data and applies clustering techniques to estimate inter-client dataset similarity. 
Based on these similarity estimates, the framework implements a client selection strategy that enables each client to collaborate with others who have compatible data distributions.
We evaluate PQFed on two benchmark datasets, CIFAR-10 and MNIST, integrated with 3 existing federated learning algorithms.
Experimental results show that PQFed consistently improves the target client's model performance, even with a limited number of participants.
We further benchmark PQFed against a baseline cluster-based algorithm, IFCA, and observe that PQFed also achieves better performance in low-participation scenarios.
These findings highlight PQFed’s scalability and effectiveness in personalized federated learning settings.
\end{abstract}


%
\IEEEpeerreviewmaketitle

\section{Introduction}
\label{sec:intro}

During the rapid development of AI technologies, machine learning models have been widely applied across various domains.
Traditional machine learning approaches typically rely on centralizing large volumes of data, which raises significant concerns regarding privacy, data security, and maintenance costs.
Federated Learning (FL), a decentralized approach to collaboratively train models without sharing raw data, offers a promising solution to these issues \cite{mcmahan2017communication}.
However, a major challenge in FL is data heterogeneity, where clients typically possess data with differing distributions. 
Previous works~\cite{DBLP:journals/corr/abs-2102-02079, ye2023heterogeneous} have shown that such distributional differences among client datasets can significantly degrade the performance of the FL global model.

To address this issue, an increasing number of studies have proposed various solutions.
One popular direction focuses on optimizing the objective function of the model by incorporating terms that account for each client's data characteristics~\cite{li2020fedprox, ye2023heterogeneous, acar2021federatedlearningbaseddynamic}.
Other approaches emphasize personalized model tuning, where clients locally adapt the global model to better fit their own local data~\cite{fallah2020personalizedfederatedlearningmetalearning, hu2020personalized}. 
Based on this, some methods propose evaluating and weighting other clients’ model updates using statistical similarity metrics to enhance local personalization~\cite{LiSample-level, shi2023makelandscapeflatterdifferentially}.
Furthermore, some research explores client clustering strategies, where clients with similar data distributions are grouped during training~\cite{xuSAFE, sattler2019clusteredfederatedlearningmodelagnostic, ghosh2020efficient}.

Although these methods alleviate the data heterogeneity problem, most existing approaches still assume full participation from all clients during training. 
This assumption leads to high computational and communication overhead as the number of training rounds increases, even though a single client often does not need collaboration with the entire group to achieve strong personalized performance.
In many real-world applications, one client may has a dominant interest in optimizing the model for their own local data. 
We refer to such a client as the target client. 
For instance, in materials science, a research lab may only need a model that accurately predicts the microstructure of its own alloy samples under specific heat treatment conditions, without generalizing to other labs or experimental setups~\cite{yue2024phase}. 
Likewise, a hospital may focus on rare diseases prevalent in its patient population~\cite{Dervishi2023}. 
In such scenarios, clients typically prioritize personalized model performance over generalization, and may be reluctant to collaborate with a large number of other parties due to privacy concerns or resource constraints.

This motivates our focus on target-client personalization under privacy constraints.
In this work, we propose PQFed, a novel privacy-preserving framework that performs early-stage quality control to evaluate client compatibility prior to federated training. 
This framework supports the target client to collaborate with a small subset of clients whose data distributions closely align with their own, thereby allows them to avoid participating in large-scale federated training.
PQFed distributes a shared, server-trained feature extractor to clients, allowing them to project their local data into a unified semantic space. 
The server then uses a clustering-based similarity assessment to recommend collaboration groups for each client to improve personalization.
To enhance privacy, we establish a privacy-preserving environment by implementing local differential privacy (LDP) on each client, where Laplacian noise is added to the extracted features before they are shared with the server. 
Unlike traditional methods, PQFed does not need to be embedded into the federated training process. 
By filtering out overly heterogeneous clients at the beginning, it conserves training resources for the remaining clients.
We also analyze the robustness of the server-side similarity estimation process under the noise introduced by LDP.


PQFed is modular and flexible, allowing easy integration with existing FL algorithms.
We evaluate PQFed on two well-known datasets, MNIST and CIFAR-10, integrating 3 different FL algorithms, FedAvg~\cite{mcmahan2017communication}, FedProx~\cite{li2020fedprox} and FedDyn~\cite{acar2021federatedlearningbaseddynamic}. 
In addition, we also compare PQFed with a benchmark clustering-based FL algorithm IFCA~\cite{ghosh2020efficient}, to highlight the differences between PQFed’s collaboration strategy and traditional clustering approaches in FL.
A series of experiments demonstrate that PQFed can accurately track the distribution distance between clients' datasets under privacy-preserving constraints. 
Building on this, the collaborative client selection strategy tailored to each client consistently enhances the global model's performance, even with a limited number of participating clients across all integrated FL algorithms, and outperforms IFCA in most cases with fewer training resources.


The remainder of this paper is organized as follows. 
Section~\ref{sec:related_work} and Section~\ref{sec:background} provide an overview of related work on FL, particularly focusing on solutions to data heterogeneity, and the necessary background for this paper. 
In Section~\ref{sec:framework}, we introduce our proposed framework, PQFed, along with the associated threat model. 
Section~\ref{sec:results} presents experimental evaluations of the proposed framework under various scenarios, including an analysis of the trade-off between added noise and computed data similarity. Finally, Section~\ref{sec:dis} offers a detailed discussion of our empirical results and outlines potential optimization directions for future work.

\section{Related Work}\label{sec:related_work}

    A lot of work in FL focuses on addressing the challenge of data heterogeneity.  
    Some research focuses on optimizing the training process based on client contributions~\cite{DBLP:journals/corr/abs-2102-02079}. 
    For instance, 
    FedBalancer \cite{shin2022fedbalancerdatapacecontrol} dynamically adjusts the training step and sample selection using sample loss values.
    FedCCEA \cite{shyn2021fedccea}, estimates client contributions by constructing an accuracy approximation model based on sampled data sizes.
    FedBR \cite{guo2023fedbrimprovingfederatedlearning} tackles the issue of biased local learning by introducing mechanisms to reduce bias in local classifiers and align local features with global ones. 
    Building on these ideas, a line of research emphasizes personalization in FL, aiming to tailor the global model to individual client distributions. 
    Fallah et al. \cite{fallah2020personalizedfederatedlearningmetalearning} propose Per-FedAvg, a meta-learning approach that enables fast adaptation to client-specific tasks by learning a shared initialization point, which allows clients to achieve personalized models through a few local gradient descent steps. 
    Similarly, Hu \cite{hu2020personalized} et al. explore methods for rapid adaptation through personalized initialization strategies or hybrid model architectures that balance shared and local components.
    To further advance personalization, some research focuses on leveraging data quality to identify high-contributing clients. 
    For example, Li et al. \cite{LiSample-level} quantify sample-level data quality using multiple metrics to guide aggregation decisions. 
    FedSDG-FS \cite{li2023fedsdgfsefficientsecurefeature} securely selecting informative features using Gaussian stochastic dual-gates in federated settings.
    Clustering-based approaches have also been explored for assessing data quality. 
    For example,
    Ghost et al. introduce IFCA~\cite{ghosh2020efficient}, which addresses client heterogeneity by assuming that clients can be partitioned into clusters, each associated with a distinct model. 
    Furthermore, the work of Tan et al. \cite{Reputation-Aware} assigns clients to low, medium, or high reputation levels based on the quality of their gradient updates, using this reputation to guide the client selection process.
    Unlike these approaches that still require all clients to participate in the training process, our method operates independently of the federated training loop and introduces an early-stage quality control mechanism to address data heterogeneity for personalized clients.

\section{Background}
\label{sec:background}

    \subsection{Federated Learning Framework}
    \label{sec:flf}

    The basic architecture of FL involves a central server coordinating training across multiple clients, denoted as \( P_1, P_2, \ldots, P_n \). 
    Each client holds a local dataset \( D_1, D_2, \ldots, D_n \), and the goal is to collaboratively train a global model by leveraging knowledge from all clients—without sharing raw data~\cite{li2020federated}.
    In this work, we evaluate our proposed method with 3 well-known FL algorithms: FedAvg, FedProx, and FedDyn. \\

    \noindent\textbf{FedAvg:}
    FedAvg~\cite{mcmahan2017communication} is the most classical FL algorithm. 
    It operates iteratively in a few steps: each client trains a local model on its dataset for several epochs, sends the resulting model parameters to a central server, where the server aggregates them via weighted averaging, and then broadcasts the updated global model back to all clients.
    This iterative process continues until convergence. Formally, the global update at round \( t \) is given by
    
    \begin{equation}
        w^{t+1} = \sum_{k=1}^K \frac{n_k}{n} w_k^{t+1},
    \label{eq:fedavg}
    \end{equation}
    where \( w_k^{t+1} \) is the updated model from client \( k \), \( n_k \) is the size of client \( k \)'s dataset, and \( n = \sum_{k=1}^K n_k \) is the total data size.\\

    \noindent\textbf{FedProx:}
    While FedAvg is simple and widely adopted, it often suffers from performance degradation when client data distributions are non-IID~\cite{fallah2020personalizedfederatedlearningmetalearning, haddadpour2019convergencelocaldescentmethods}.
    To mitigate this issue, FedProx~\cite{li2020fedprox} adds a proximal term to the local objective, constraining local updates and reducing client drift. 
    The optimization problem for client \( k \) becomes:

    \begin{equation}
        \min_{w} \ f_k(w) + \frac{\mu}{2} \| w - w^t \|^2,
    \label{eq:fedprox}
    \end{equation}
    where \( f_k(w) \) is the local loss function, \( w^t \) is the global model at round \( t \), and \( \mu \) is a coefficient that controls the strength of the regularization. 
    This additional term encourages local updates to remain close to the global model, improving stability in heterogeneous settings. \\
    
    \noindent\textbf{FedDyn:}
    Similar to FedProx, FedDyn~\cite{acar2021federatedlearningbaseddynamic} introduces a dynamic regularization mechanism that aligns each client's objective with the global direction. 
    The local objective at round \( t \) is defined as:

    \begin{equation}
        \min_{w} \ f_k(w) - \langle h_k^t, w \rangle + \frac{\lambda}{2} \| w \|^2,
    \label{eq:feddyn}
    \end{equation}
    where \( h_k^t \) is a dynamic control variate that tracks the global model's trajectory, and \( \lambda \) is a regularization parameter. 
    This dynamic adjustment helps FedDyn better align local updates with the global objective, leading to improved convergence and robustness in heterogeneous data settings.\\

    \noindent\textbf{IFCA:}
    The Federated Clustering Algorithm (IFCA)~\cite{ghosh2020efficient} tackles data heterogeneity by clustering clients and associating each cluster with its own model.
    Rather than maintaining a single global model, the server keeps \( k \) separate models throughout training.
    In each communication round, clients evaluate all \( k \) models locally and identify the one that yields the lowest loss on their data.
    Each client then performs local updates on the selected model and sends the result back to the server.
    The server aggregates updates cluster-wise, refining each model based on feedback from its associated clients.
    This strategy allows IFCA to learn multiple specialized models personalized to different client groups.


    \subsection{Principal Component Analysis}
    \label{sec:pca}

    In this work, we use Principal Component Analysis (PCA) \cite{abdi2010principal} as the feature extraction in our framework.
    PCA reduces the dimensionality of high-dimensional data while preserving its most significant variations. 
    PCA computes new orthogonal variables—called principal components—that are linear combinations of the original features. 
    These components are derived from the eigenvectors and eigenvalues of the dataset’s covariance matrix, ordered by decreasing eigenvalues. The contribution of each component is quantified by the \textit{Explained Variance Ratio}, which indicates the proportion of total variance captured.
    Recent studies have demonstrated PCA’s effectiveness in privacy-preserving data communication. 
    For example, Froelicher et al. \cite{sppca} show that PCA-based encoding can achieve accuracy comparable to non-secure centralized solutions, regardless of how the data is distributed across clients. 
    Furthermore, distributed PCA techniques have been proposed specifically for federated settings, where each client computes a local covariance matrix and securely aggregates it on the server to build a global PCA model \cite{dispca}.

    \subsection{K-means Clustering and Distance Between Datasets}\label{sec:k-means}
    We use K-means~\cite{lloyd1982least} as the clustering model in our framework. K-means is a widely used method for partitioning unlabeled data into \( K \) clusters. The algorithm begins by randomly initializing \( K \) centroids and then iteratively assigns each data point to the nearest centroid. After assigning, the centroids are updated based on the new cluster memberships. This process repeats until the centroids converge and stabilize. The number of clusters \( K \) is a tunable parameter, allowing the model to adapt to different expected client groupings.
    The cluster centers obtained from K-means can be directly used to compute distances between datasets. 
    To extend this comparison from individual clusters to entire groups of clusters, Earth Mover’s Distance (EMD)~\cite{rubner2000earth}, a metric grounded in optimal transport theory, provides a powerful tool. 
    EMD calculates the minimal cost required to transform one probability distribution into another, capturing both the spatial structure and distributional differences between datasets. It is defined as:
        \begin{equation}
        \text{EMD}(P, Q) = \min_{F} \sum_{i,j} f_{ij} \cdot d(x_i, y_j),
        \end{equation}
    where \( P = \{(x_i, p_i)\} \) and \( Q = \{(y_j, q_j)\} \) are two discrete probability distributions, with \( x_i \) and \( y_j \) representing support points and \( p_i \), \( q_j \) their associated weights. The term \( f_{ij} \) denotes the optimal flow from point \( x_i \) in \( P \) to point \( y_j \) in \( Q \), and \( d(x_i, y_j) \) is the ground distance (e.g., Euclidean) between those points. 
    The EMD seeks the flow configuration \( F \) that minimizes the total transport cost, thus providing a meaningful measure of dissimilarity between distributions.
    Recent work from Alvarez-Melis  et al.~\cite{NEURIPS2020_f52a7b26}, also demonstrates how EMD can be effectively used to measure dataset similarity in a geometrically meaningful way. 
    This approach treats datasets as empirical distributions embedded in a feature space and leverages optimal transport to quantify their structural differences. 

    \subsection{Membership Inference Attack}
    A membership inference attack (MIA) seeks to determine whether a specific data record was part of a model’s training set. By querying the model (for example via its confidence scores, loss values, or gradient updates), an adversary can exploit differences in the model’s behavior on seen versus unseen examples to infer membership, often due to overfitting or memorization. Formally, given a target sample \(x\) and black-box (or white-box) access to a trained model \(f\), the attacker’s goal is to decide between the hypotheses
    \[
      H_{\mathrm{in}}: x \in \mathcal{D}_{\mathrm{train}}
      \quad\text{versus}\quad
      H_{\mathrm{out}}: x \notin \mathcal{D}_{\mathrm{train}}.
    \]
    Successful MIAs undermine training data privacy, since correct membership decisions leak sensitive information about individuals. Previous work has shown that models with high generalization gaps or those trained on small datasets are especially vulnerable, motivating the use of differential privacy and other defenses to mitigate membership leakage~\cite{Yeom2017-ho, Shokri2017-ps, Carlini2022-nw}.

    \subsection{Local Differential Privacy} 
    \label{sec:ldp}
    Local Differential Privacy (LDP)~\cite{truex2020ldp} is a variant of Differential Privacy(DP), designed to ensure that each user can anonymize their data locally before sharing it. 
    Unlike DP, which typically requires a centralized trusted client to have access to the data before analysis, LDP eliminates the need for such a trusted client.
    Formally, an algorithm \( A \) satisfies \(\epsilon\)-LDP if, for any possible inputs \( x \) and \( x' \), and any output \( y \), the following condition is met:
    \begin{equation}
        \text{Pr}[A(x) = y] \leq e^{\epsilon} \text{Pr}[A(x') = y]
    \end{equation}
    This condition ensures that the probability of producing any output does not differ significantly when the input is changed, thereby preserving privacy.
    Furthermore, for a numerical function \( f(x) \), the function \( F(x) \) satisfies \(\epsilon\)-LDP when Laplacian noise is added as follows~\cite{huang2020improving}:
    \begin{equation}
    F(x) = f(x) + \text{Lap}\left(\frac{s}{\epsilon}\right)
    \label{equ:lap}
    \end{equation} 
    where \( s \) represents the sensitivity of the function \( f \), which quantifies the maximum change in the function's output due to a single input change.
    The Laplace distribution is characterized by the following probability density function:
    \begin{equation}
           f(x | \mu, \lambda) = \frac{1}{2\lambda} \exp\left(-\frac{|x - \mu|}{\lambda}\right)
    \end{equation}
    where $\lambda = \frac{s}{\epsilon}$ and $\mu$ is a local parameter. 
    In this work, we apply \(\epsilon\)-LDP to the results of PCA by adding Laplacian noise.
    Specifically, we used the \( l_1 \) sensitivity, defined as:
    \begin{equation}
    s = \max ||\text{f(x)} - \text{f(x')}||
    \end{equation}

    We specifically apply LDP directly to data transformed via PCA before sharing with the server. 
    Similar approaches have been explored to provide privacy guarantees for dimension-reduced data in collaborative scenarios. 
    For example, Jiang et al.~\cite{jiang2021federated} demonstrated the effectiveness of applying differential privacy techniques to PCA components to mitigate inference risks while preserving data utility. 
    Moreover, Li et al.~\cite{li2020federated} validated that introducing local differential privacy mechanisms into intermediate representations, such as PCA features, further strengthens resistance to membership inference attacks by ensuring the indistinguishability of individual data records.

\section{Proposed Framework}
\label{sec:framework}

    \subsection{System Model}
    In many real-world applications of federated learning, some clients have a dominant interest in optimizing model performance specifically on their own local data. 
    We refer to such a participant as a target client.
    For example, a research lab may only need a model that works well for its own materials, like certain alloys or processing conditions, instead of all data from every participant. 
    Similarly, a hospital may focus on accurate predictions for rare diseases common in its own patients, rather than overall performance on unrelated cases.
    In such cases, clients are often constrained by privacy concerns, regulatory restrictions, or limited computational and communication resources. 
    These constraints reduce their willingness to collaborate with a large number of other participants. 
    As a result, target clients usually prefer to work with just a few collaborators who have similar data, rather than taking part in full federated training.
    Based on this, our system model consists of two main entities: multiple clients, including the target client \( T \), and a central server \( S \). 
    Clients may have data from diverse classes, leading to non-IID (Independent and Identically Distributed) data. 
    As a result, each client's dataset exhibits distinct statistical properties. 
    In this work, we assume the target client \( T \) has a local train dataset \( \mathcal{D}^{\text{train}}_T \) and a local testing dataset \( \mathcal{D}^{\text{test}}_T \), drawn from a similar distribution. 
    To achieve a highly accurate personalized model, client \( T \) participates in FL by collaborating with only a limited subset of other clients.
    Under this constraint, \( T \) aims to identify suitable clients among the available set (denoted as \( P_1, P_2, \dots, P_n \)) that share a similar dataset to maximize its learning benefits based on its testing dataset, \( \mathcal{D}^{\text{test}}_T \). 
    In order to align with the principles of FL, this method must ensure decentralization by avoiding the exchange of raw data while maintaining privacy-preserving feature representation sharing.
    
    \subsection{Threat Model}
    In our proposed scheme, we adopt an honest-but-curious threat model. 
    The central server faithfully executes the prescribed protocols but may attempt to infer sensitive information from shared data. In particular, we consider the central server as a potential adversary capable of performing membership inference attacks (MIA). To avoid centralized collection of raw data, we leverage the FL paradigm, sharing only feature representations of each client’s dataset obtained via PCA. This dimensionality reduction inherently reduces data details and provides a preliminary layer of privacy protection. Nonetheless, MIA remains a pervasive threat~\cite{Yeom2017-ho, Shokri2017-ps, Carlini2022-nw}, especially when models overfit or when insufficient noise is applied. To address this, our framework injects Laplacian noise calibrated to each component’s sensitivity and the privacy parameter $\epsilon$, thereby balancing privacy and utility. We evaluate the residual privacy risk by simulating a Euclidean distance–based membership inference attack, inspired by prior works~\cite{Dervishi2023, Halimi2022}. Further methodological details and empirical results are provided in Section~\ref{sec:privacy_analysis}.

    \subsection{PQFed: A Privacy-Preserving Quality-Controlled Federated Learning Framework}
    
\begin{figure*}[h!]
      \centering
      \includegraphics[width=\linewidth]{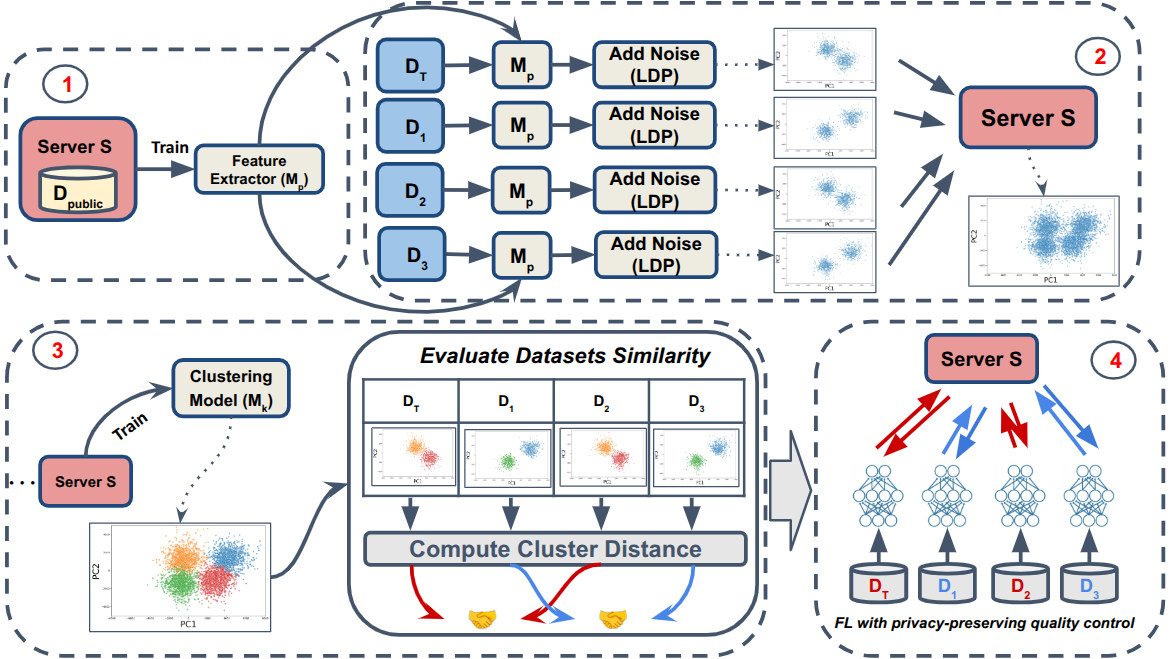}
      \caption{Overview of PQFed. (1) A feature extractor (e.g., PCA), \( M_{public} \), is trained on a public dataset \( D_p \) at the server and sent to clients. (2) Clients use \( M_p \) to extract features from their local datasets, apply LDP by adding Laplace noise, and send the noisy representations to the server. (3) The server trains a clustering model \( M_k \) (e.g., K-means) using the received features and evaluates data similarity by computing cluster distances using the Earth Mover's Distance. (4) Based on the similarity metrics, the server performs federated aggregation separately for groups of clients with similar dataset distributions. Solid arrows represent actual data flow, while dashed arrows indicate visualization of the current data state.}
      \label{fig:QC}
    \end{figure*}

    PQFed is designed to align with the FL framework, ensuring seamless integration while maintaining privacy-preserving properties.
    The workflow is illustrated in Figure~\ref{fig:QC}.
    To demonstrate the process, we consider a scenario with $n$ clients; however, the approach can be easily extended to accommodate a larger number of clients.

    To provide a highly accurate personalized model for the target client $T$ via FL, it is important to compare the dataset of $T$ with those of other clients to identify suitable collaborators. 
    However, due to privacy concerns, directly sharing raw datasets with the central server is not feasible. 
    To address this, we propose exchanging feature representations of each client’s data instead of the raw data.
    In our work, we utilize PCA for feature extraction. 
    The central server \( S \) is responsible for providing a PCA model, $M_p$, as an anchor to project all clients' datasets into a unified feature space.
    The PCA model will be trained on a public dataset $D_{public}$.
    Once the model \( M_p \) is trained, it is broadcast to all clients. 
    Each client \( P_i \) applies \( M_p \) to its local dataset \( D_i \) and obtains a low-dimensional representation. 
    To ensure privacy, each client then applies LDP by adding Laplacian noise controlled by a privacy budget, $\epsilon$, to the PCA results.
    These noisy PCA representations are sent back to the server \( S \).
    Finally, after receiving all the noisy representations from the clients, server \( S \) aggregates them and applies a K-means model $M_k$ to obtian the distribution of each client’s data across the identified clusters.
    To measure the similarity between the target client \( T \) and the other clients, the server $S$ computes the Earth Mover's Distance (EMD) between their respective clustering distributions.
    
    Based on the computed EMD values, the server can manually select a threshold range from 30\% to 60\% of the maximum observed EMD value to categorize clients into three groups (as illustrated in Figure~\ref{fig:emd_thresholds}). 
    A strict selection applies a 30\% threshold, excluding any clients whose distance exceeds this limit, thereby including only those highly similar to the target client.
    A lenient selection adopts a 60\% threshold, allowing a broader range of clients to participate; 
    Clients with distance above 60\% are excluded from collaboration.
    After selecting the clients, multiple FL algorithms such as FedAvg, FedProx, and FedDyn can be applied using only the identified clients to build the personalized model for client \(T\). 
    We summarize the steps of PQFed in Algorithm~\ref{alg:pqfed}.

    \begin{algorithm}[htbp]
        \caption{The PQFed Framework}
        \label{alg:pqfed}
        \KwIn{Target client $T$, clients $\{P_1, \ldots, P_n\}$, local dataset $D_i$ at each client $P_i$, public dataset $D_{\text{public}}$}
        \KwOut{Recommended collaborators for client $T$}
        
        \textbf{Server $S$ initiates:}\\
        Train feature extractor $M_p$ using $D_{\text{public}}$ \\
        Distribute $M_p$ to all clients $\{P_1, \ldots, P_n\}$ \\
        
        \textbf{Each client $P_i$ executes:}\\
        Apply $M_p$ to extract features: $Z_i \leftarrow M_p(D_i)$ \\
        Apply LDP: $\tilde{Z}_i \leftarrow Z_i + \mathcal{L}(\epsilon)$ \\
        Send $\tilde{Z}_i$ to server $S$ \\
        
        \textbf{Server $S$ executes:}\\
        Aggregate received representations $\{\tilde{Z}_i\}_{i=1}^n$ \\
        Train clustering model: $\mathcal{C} \leftarrow M_k(\{\tilde{Z}_i\})$ \\
        For each client $P_i$, compute Earth Mover's Distance based on $\mathcal{C}$: $d_i = \text{EMD}(\tilde{Z}_T, \tilde{Z}_i)$ \\
        Determine EMD threshold $\tau$ manually \\
        Return clients $\{P_i \mid d_i \leq \tau\}$ as recommended collaborators  
    \end{algorithm}

\section{Evaluation}\label{sec:results}

    \subsection{Datasets and Experimental Setup}
    \label{sec:dataset}

    We conduct our experiments on two widely-used image classification benchmarks: CIFAR-10~\cite{krizhevsky2009learning} and MNIST~\cite{mnist}, both consisting of 10 classes. 
    Each dataset contains a training set, \( \mathcal{D}^{\text{train}}_{Global}\) and a test set, \( \mathcal{D}^{\text{test}}_{Global}\).
   
    To simulate realistic federated settings with non-IID client distributions, we apply the same strategy to construct the target client \( T \) on both CIFAR-10 and MNIST, following these steps:
    
    \begin{enumerate}
        \item \textbf{Public PCA Construction:}  
        We randomly select 500 images from the
        \( \mathcal{D}^{\text{test}}_{Global}\) to construct a small public dataset. 
        This subset is used to train a PCA model \( M_{\mathrm{Public}} \) with 50 components (determined via the elbow method).
        
        \item \textbf{PCA Feature Extraction and Clustering:}  
        We apply the trained PCA model to transform the whole training set, \( \mathcal{D}^{\text{train}}_{Global}\). 
        We then perform K-means clustering with \( K = 15 \) clusters (also chosen via the elbow method) on these PCA-transformed data. 
        These clusters serve as proxies for distinct data distributions, simulating natural heterogeneity among clients.
        
        \item \textbf{Client \( T \) Construction:}  
        We construct the target client \( T \) by sampling data from distinct clusters.  
        In particular, we randomly select 3 clusters, denoted as \( \mathcal{C}_{\mathrm{T}} = \{c_i, c_j, c_k\} \), and sample 400 images from each to form the training dataset, \( \mathcal{D}^{\text{train}}_T\) for the client \( T \).  
        For the test set of client \( T \), \( \mathcal{D}^{\text{test}}_T\) , we sample images from \( \mathcal{C}_{\mathrm{T}}\) in the same proportions as each cluster’s distribution in the training set.  
        We use 300 images for CIFAR-10 and 1,200 images for MNIST as the test set respectively, the latter being larger due to the dataset’s lower complexity.
    \end{enumerate}
    This setup allows for meaningful personalization: other clients are constructed from either the same clusters (for similar distributions) or different clusters (for dissimilar ones), enabling evaluation of the model's ability to identify compatible collaborators, while also ensuring the training and test sets of client $T$ share the same data distribution.
    
\begin{figure}[h]
    \centering
    \begin{subfigure}[b]{0.23\textwidth}
        \centering
        \includegraphics[width=\textwidth]{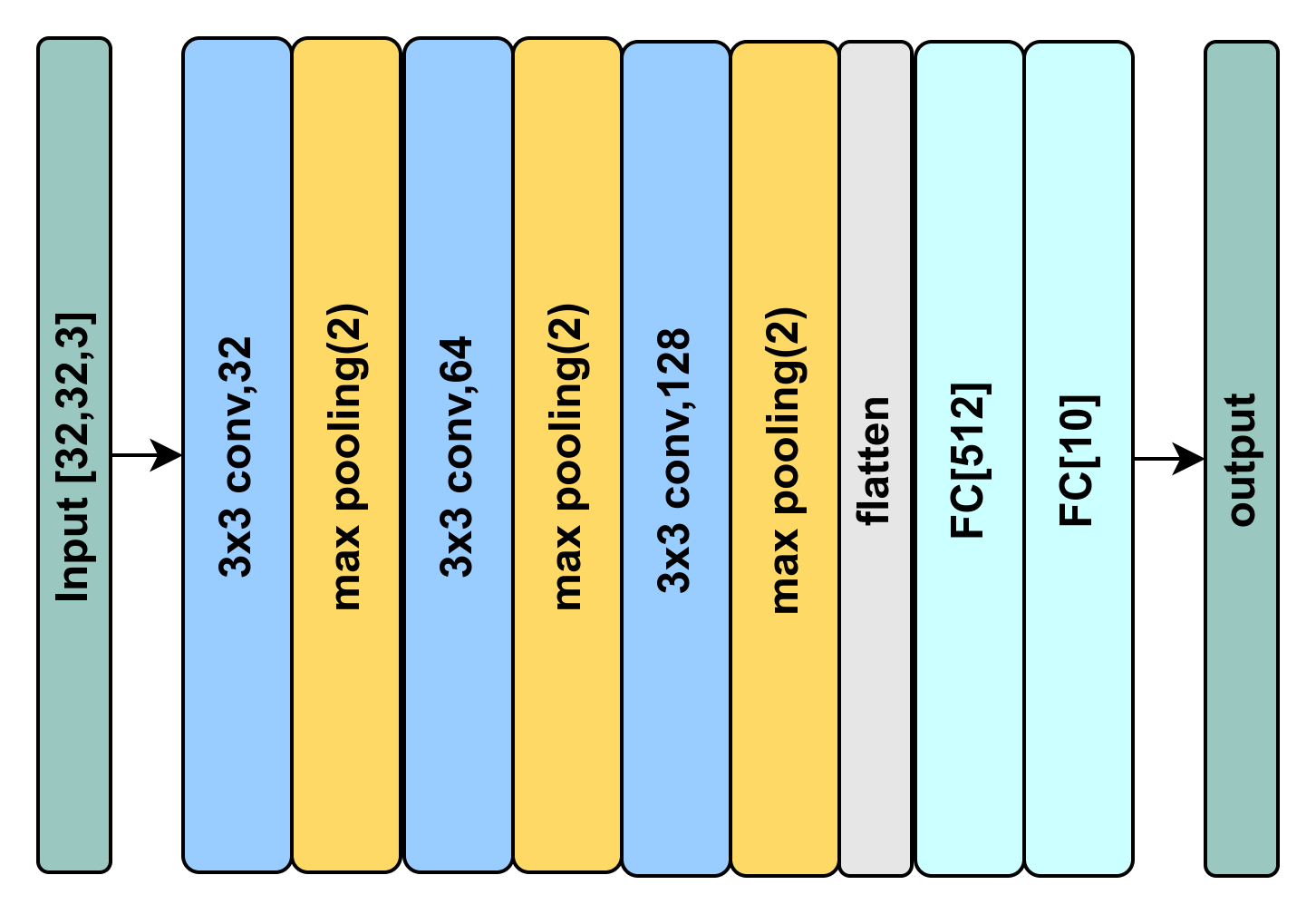}
        \caption{CNN Architecture for CIFAR-10}
        \label{fig:cnn_architecture_cifar}
    \end{subfigure}
    \hfill
    \begin{subfigure}[b]{0.23\textwidth}
        \centering
        \includegraphics[width=\textwidth]{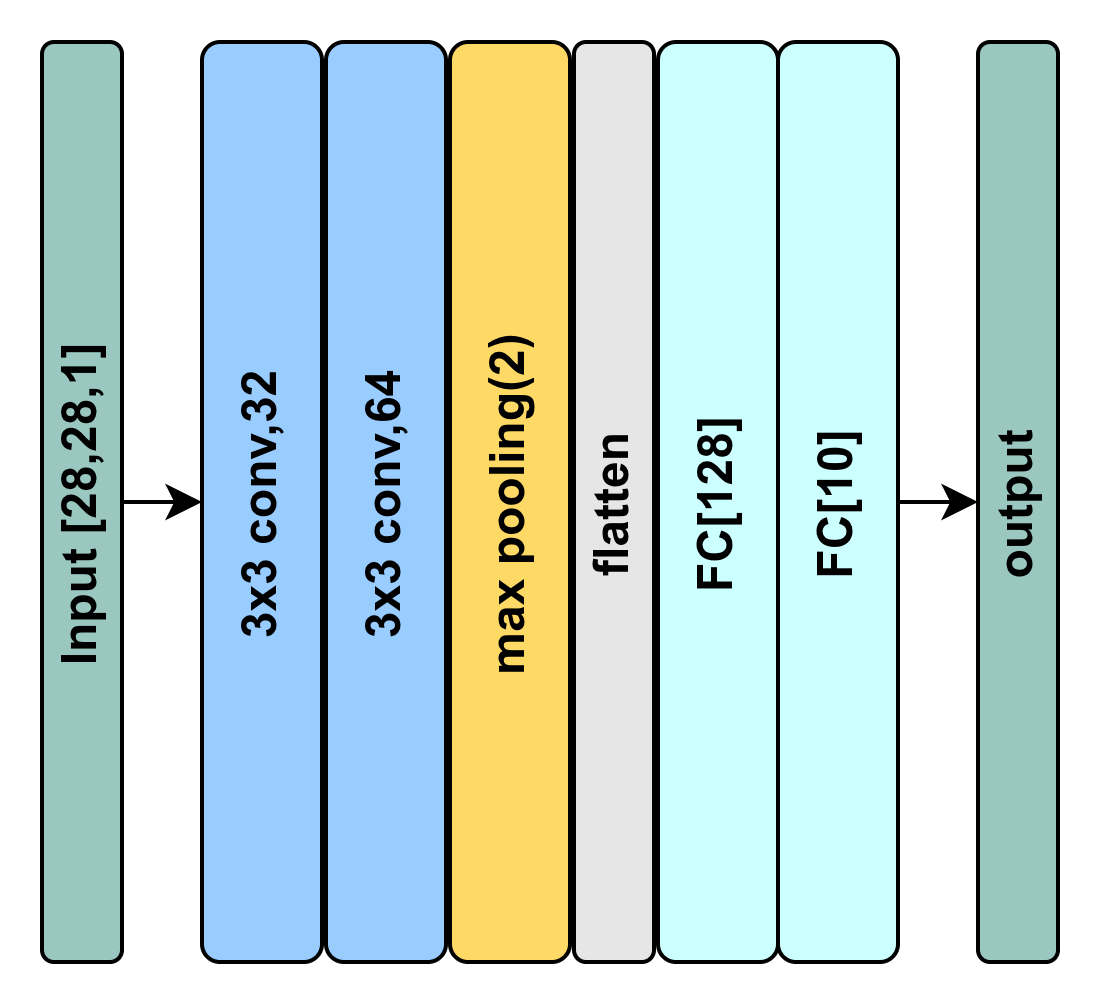}
        \caption{CNN Architecture for MNIST}
        \label{fig:cnn_architecture_mnist}
    \end{subfigure}
    \caption{CNN architectures for CIFAR-10 and MNIST datasets. 
(a) For CIFAR-10 model, each image has input size of \([32, 32, 3]\), and includes 2 convolutional layers followed by one max pooling layer and 2 fully connected layers. 
(b) For MNIST model, each image has input size of \([28, 28, 1]\), and consists of 2 convolutional layers followed by one max pooling layer and 2 fully connected layers.}

    \label{fig:cnn_arch}
    \end{figure}

    We employ CNN models as the global model,
    with the architecture illustrated in Figure~\ref{fig:cnn_arch}. 
    Training is conducted in a PyTorch environment on a single V100 GPU, and the training hyperparameters are detailed in Table~\ref{tab:hyperparams}. 

      \begin{table}[h]
    \centering
    \renewcommand{\arraystretch}{1.2}
    \setlength{\tabcolsep}{12pt}
    \caption{Hyperparameters for FL Experiments on CIFAR-10 and MNIST.}

    \begin{tabular}{lcccc}
        \hline
        \textbf{Hyperparameter} & \textbf{CIFAR-10} & \textbf{MNIST} \\
        \hline
        Learning Rate & $1 \times 10^{-3}$ & $1 \times 10^{-3}$ \\
        Batch Size & 32 & 16 \\
        Epochs per Round & 1 & 1 \\
        Total Rounds & 20 & 20 \\
        \hline
    \end{tabular}
    \label{tab:hyperparams}
    \end{table}

    \subsection{Quantifying the Dissimilarity Between Clients' Datasets}
    \label{sec:dissim-rate}
    To better evaluate the significance of PQFed, it is important to employ a metric that quantifies the dissimilarity between clients.
    As described in the section~\ref{sec:dataset}, 
    client $T$ is formed by sampling 400 points each from 3 clusters in $\mathcal{C}_{\mathrm{T}}$.
    To observe different levels of similarity between target client's and other clients' datasets, we define a dissimilarity rate \( r \in [0,1] \) to construct the datasets of other clients. 
    This rate \( r \) controls the proportion of data that is used from $\mathcal{C}_{\mathrm{T}}$ and a different, randomly selected cluster \( c_{\mathrm{diff}} \) in order to form the dataset of a client. 
    For a given rate $r$, each client is constructed by sampling
    $(1 - r) \cdot 400 $ points from each cluster in $\mathcal{C}_{\mathrm{T}}$ and $r \cdot 400$ points from $c_{\mathrm{diff}}$.
    For instance, $r=0$ produces a client identical to $T$, while $r=1$ yields a client  sampling data from all clusters except for the ones in $\mathcal{C}_{\mathrm{T}}$. 
    To observe a linear trend with increasing dissimilarity,
    we construct 10 clients with \( r \) values from 0.1 to 1 in steps of 0.1 (i.e., \( r = 0.1, 0.1, 0.2, \ldots, 1.0 \)).
    After forming the datasets of all clients this way, we quantify the dissimilarity for these clients using the Earth Mover’s Distance (EMD) between their datasets.  
    The EMD is computed based on the K-means clustering results of their PCA-transformed data, using the public PCA model $M_{Public}$.

\begin{figure}[htbp]
        \centering
        \subfloat[CIFAR-10]{\includegraphics[width=0.45\textwidth]{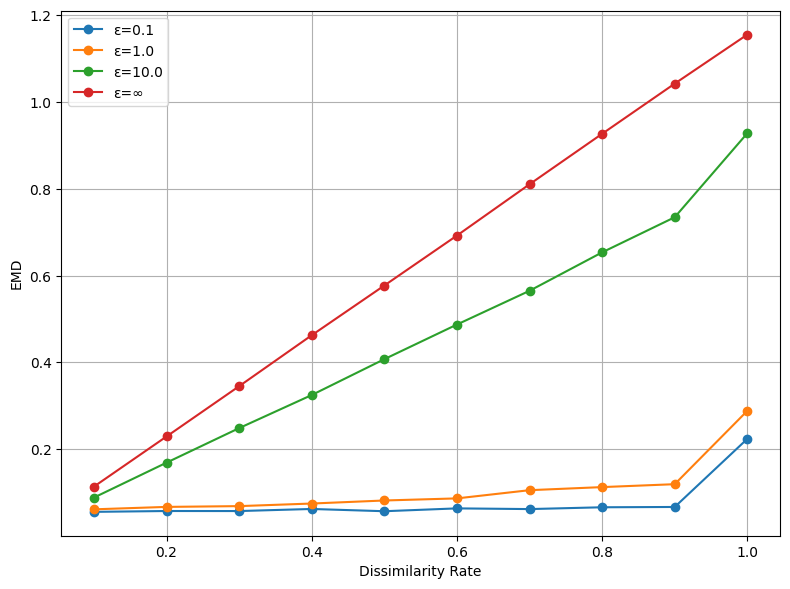}\label{fig:cifar10_all}}
        \\
        \subfloat[MNIST]{\includegraphics[width=0.45\textwidth]{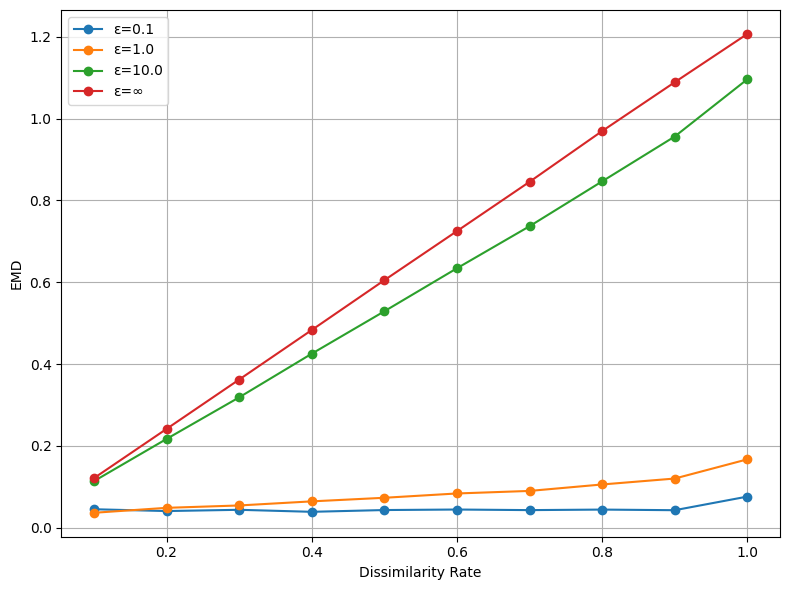}\label{fig:mnist_all}}
        \caption{Earth Mover's Distance under varying data dissimilarity rates with and without differential privacy noise ($\epsilon=\{0.1, 1, 10, \infty\}$). 
        }
        \label{fig:emd_results}
    \end{figure}

    Since each client needs to apply LDP to their PCA-transformed data to protect privacy, we also examine how LDP affects the EMD between these clients. 
    Specifically, we add Laplacian noise to PCA-transformed data using privacy budgets \(\epsilon \in \{0.1, 1, 10\}\).  
    The selection of \(\epsilon\) values will be discussed in Section~\ref{sec:privacy_analysis}.
    We repeated this process 50 times and show average results in Figure~\ref{fig:emd_results}.

    It illustrates how the EMD varies with the dissimilarity rate (\(r\)) under different privacy budgets on CIFAR-10 (Figure~\ref{fig:cifar10_all}) and MNIST (Figure~\ref{fig:mnist_all}).
    Under the no-noise baseline, shown as red line (\(\epsilon=\infty\)), the EMD increases roughly linearly from about 0.12 at \(r=0.1\) to 1.16 (CIFAR-10) or 1.20 (MNIST) at \(r=1.0\), reflecting the growing distributional shift. 
    With moderate noise (\(\epsilon=10\)), the curves closely track the baseline, rising to 0.93 and 1.10 respectively, indicating minimal reduction. 
    At \(\epsilon=1\), EMD still grows but more slowly, approximately 0.06 to 0.28 on CIFAR-10 and 0.04 to 0.17 on MNIST, which shows partial masking of the shift. 
    Finally, strong noise (\(\epsilon=0.1\)) flattens the curve (EMD $<0.08$ for all \(r<1.0\)), demonstrating that heavy Laplace perturbation effectively hides even large heterogeneity (the impact by strong noise on the EMD are further discussed in Section~\ref{sec:dp_on_emd}). 

    \subsection{Validation of PQFed Similarity Schema}
    \label{sec:1to5}
    To test the effectiveness of PQFed similarity-based schema, we evaluate whether clients defined as similar can help improve FL global model performance on client $T$’s testing set, more than dissimilar ones.
    As described in Section~\ref{sec:dissim-rate}, we construct similar clients using a dissimilarity rate of \( r = 0 \), and dissimilar datasets using \( r = 1 \). 
    We then conduct 5 experiments under different scenarios, varying the number of similar and dissimilar participating clients in FL.
    These experiments were carried out on both the MNIST and CIFAR-10 datasets.
    The settings for each scenario are described below, and the corresponding results are reported in Table~\ref{tab:1t5results}.

     \begin{itemize}
            \item In \textbf{Experiment 1}, local model is trained using only client \( T \)'s training dataset.
        
            \item In \textbf{Experiment 2}, A FL global model is trained with client \( T \) and 4 other similar clients.
        
            \item In \textbf{Experiment 3}, A FL global model is trained with client \( T \) and 4 other dissimilar clients.
        
            \item In \textbf{Experiment 4}, A FL global model is trained with client \( T \) and 9 other clients, where 4 clients are similar and 5 are dissimilar.
        
            \item In \textbf{Experiment 5}, A FL global model is trained with client \( T \) and 19 other clients, where 4 clients are similar and 15 are dissimilar.
    \end{itemize}
         
    \begin{table*}[h]
        \centering
        \renewcommand{\arraystretch}{1.2}
        \setlength{\tabcolsep}{9pt}
        \caption{Experiments on CIFAR-10 and MNIST}
        \label{tab:1t5results}
        \begin{tabular}{p{3cm} p{5cm} p{4cm} p{3cm}}
            \hline
            \textbf{Exp.} & \textbf{Training Setup} & \textbf{CIFAR-10 Accuracy (\%)} & \textbf{MNIST Accuracy (\%)} \\
            \hline
            1 & $T$ local training only & 46.67 & 95.44 \\
            2 & FL: $T$ + 4 similar clients & \textbf{72.33} & \textbf{98.56} \\
            3 & FL: $T$ + 4 dissimilar clients & 51.67 & 96.00 \\
            4 & FL: $T$ + 4 similar + 5 dissimilar clients & 63.00 & 98.42 \\
            5 & FL: $T$ + 4 similar + 15 dissimilar clients & 60.23 & 98.06 \\  
            \hline
        \end{tabular}
    \end{table*}

    For the 5 experimental settings: Experiment 1 shows the performance of the local model, serving as the baseline. 
    Experiments 2 and 3 examine the impact of collaboration with similar and dissimilar clients, respectively, while Experiments 4 and 5 explore the effect of increasing the number of dissimilar clients. 
    From the baseline model (Experiment 1), we observe that CNN performance is suboptimal when training solely on client $T$'s local dataset, for CIFAR-10, which has complex features. 
    The local model achieves only 46.67\% accuracy, highlighting its limitations. 
    Any other FL setups yield better results compared to training in isolation. 
    As expected, when collaborating with similar distribution datasets, even with just 4 additional clients (Experiment 2), Client $T$ benefits the most, achieving 72.33\% accuracy. 
    In contrast, collaboration with 4 dissimilar clients (Experiment 3) results in a minimal performance improvement, with an accuracy of only 51.67\%.
    Experiments 4 and 5 further investigate the effect of increasing the number of similar and dissimilar clients in an FL setting. 
    In Experiment 4, despite collaborating with nine other clients (4 similar and 5 dissimilar), the global model achieves an accuracy of only 63.00\%, which is lower than the one in Experiment 2. 
    This suggests that introducing dissimilar clients introduces noise into the FL training process, limiting the performance gains for client $T$. 
    In scenarios where collaboration involves dissimilar datasets, adding more data and increasing training time does not necessarily yield the best results for the target client.
    Furthermore, in Experiment 5, where 10 additional dissimilar clients are introduced on top of Experiment 4, the FL global model's performance further declines to 60.23\%. 

    The same performance trend is observed in the MNIST dataset. 
    However, due to the strong performance of CNN on this dataset, the baseline accuracy is already quite high--95.44\% even with local training--making the performance comparisons less pronounced.  
    Nonetheless, we still see a clear improvement when client $T$ collaborates with four similar clients, reaching the highest accuracy of 98.56\%. 
    However, as more dissimilar clients are introduced, the performance gradually decreases. 
    When collaborating with 4 similar and 5 dissimilar clients, the accuracy slightly drops to 98.42\%, and further decreases to 98.06\% when 15 dissimilar clients are involved. 
    Notably, when collaborating only with dissimilar clients, the accuracy is significantly lower at 96.00\%, reinforcing the observation that collaboration with similar clients is the most beneficial for the target client.

    Our results indicate that collaboration among similar clients is more important than simply increasing the amount of data used to train the FL global model. 
    As more dissimilar datasets participate in training, the global model drifts further from client $T$’s optimal representation, leading to performance degradation.

    \subsection{Privacy Analysis}
    \label{sec:privacy_analysis}
    
    In this work, we assume an honest but curious server possesses the true PCA-transformed data of a target individual and attempts to determine their presence in the shared noisy dataset by a client. 
    To assess the privacy risk, we implement a Euclidean distance-based membership inference attack~\cite{Dervishi2023, Halimi2022}. 
    The attack leverages the Euclidean distance, $||\mathbf{x} - \mathbf{y}||_2$, to compare the target's PCA representation, $\mathbf{x}$, with those in the noisy dataset, $\mathbf{y}$.
    
    We partition the population into two distinct groups: the target group, comprising individuals whose original PCA representations are present in the noisy dataset (before noise addition), and the non-target group, consisting of individuals whose original PCA representations are excluded.
    
    

    To quantify the privacy risk associated with releasing noisy PCA‐transformed data, we employ an iterative, Euclidean‐distance‐based membership‐inference procedure. Let \(X\) denote the full set of PCA‐transformed data points stored on the server. In each iteration (or “round”) of the attack, we perform the following steps:

    First, we randomly split \(X\) into two equal‐sized subsets, \(A\) and \(B\). We treat \(A\) as the “reference” set and \(B\) as the “held‐out” set. From these two sets, we then draw up to \(n\) samples each (without replacement): let \(A_{\text{sample}}\subseteq A\) and \(B_{\text{sample}}\subseteq B\) be two subsets of size \(\min(n,\,|A|,\,|B|)\). We designate \(B_{\text{sample}}\) as the case group (points whose membership in \(A\) we wish to test) and \(A_{\text{sample}}\) as the control group (points known not to be in \(A\) for that particular round). By construction, none of the control points appear in the reference set.
    
    Next, we add Laplacian noise to the entire reference set \(A\), thereby obtaining a noisy reference set \(\widetilde{A}\). Concretely, we compute the \(\ell_{1}\) sensitivity of \(A\) coordinate‐wise:
    \[
        \mathrm{sensitivity}\bigl(A\bigr)
        \;=\; \max_{x\in A} x \;-\; \min_{x\in A} x,
    \]
    which produces a vector of length equal to the PCA‐transformed data dimension. If we denote this vector by \(\boldsymbol{s} = \bigl[s_{1},s_{2},\dots,s_{d}\bigr]\), then for each coordinate \(j\in\{1,\dots,d\}\), we draw independent Laplace noise of scale \(b_{j} = \tfrac{s_{j}}{\epsilon}\). Adding this noise coordinate‐wise to every point in \(A\) yields the perturbed set
    \begin{align*}
        \widetilde A
        &= \bigl\{\,a + \eta \;\bigm|\; a\in A,\;\eta=(\eta_1,\dots,\eta_d),
           \;\eta_j\!\sim\!\mathrm{Lap}\bigl(0,\tfrac{s_j}{\epsilon}\bigr)\bigr\}\\
        &= \bigl\{\,a + \sum_{j=1}^d \mathrm{Lap}\bigl(0,\tfrac{s_j}{\epsilon}\bigr)\,e_j
           \;\bigm|\; a\in A\bigr\}.
    \end{align*}
    where \(e_{j}\) denotes the standard basis vector in coordinate \(j\). This noise‐addition step guarantees that each coordinate of each point in \(A\) satisfies \(\epsilon\)-LDP.
    
    After obtaining \(\widetilde{A}\), we compute, for every control point \(x_{\mathrm{ctrl}}\in A_{\text{sample}}\), the Euclidean distance to each noisy reference point in \(\widetilde{A}\) and then record the minimum distance:
    \[
        d_{\mathrm{ctrl}}(x_{\mathrm{ctrl}})
        \;=\;
        \min_{\tilde{a}\in \widetilde{A}} \|\,x_{\mathrm{ctrl}} - \tilde{a}\,\|_{2}.
    \]
    Similarly, for each case point \(x_{\mathrm{case}}\in B_{\text{sample}}\), we compute
    \[
        d_{\mathrm{case}}(x_{\mathrm{case}})
        \;=\;
        \min_{\tilde{a}\in \widetilde{A}} \|\,x_{\mathrm{case}} - \tilde{a}\,\|_{2}.
    \]
    Intuitively, if a held‐out point originally belonged to \(A\), even after noise it will, on average, lie closer to \(\widetilde{A}\) than a truly external point does.
    
    We then select a threshold \(\gamma\) based on the control distances so as to fix the false‐positive rate at \(\alpha\). Concretely, we let
    \[
        \gamma 
        \;=\; 
        \text{the smallest value such that } 
        \Pr\bigl(d_{\mathrm{ctrl}} \le \gamma\bigr) 
        \;=\; \alpha.
    \]
    By construction, only an \(\alpha\)-fraction of control distances fall below \(\gamma\), thereby guaranteeing that the probability of incorrectly labeling a control point as a “member” is approximately \(\alpha\).
    
    Finally, we declare each case point \(x_{\mathrm{case}}\) to be a member of \(A\) if and only if
    \[
        d_{\mathrm{case}}(x_{\mathrm{case}})\;\le\;\gamma.
    \]
    The fraction of case points satisfying this inequality is the attack’s empirical success rate (membership inference power) for the current round.
    
    Because the above procedure involves random splitting of \(X\) into \(A\) and \(B\), random subsampling of case and control, and random draws of Laplace noise, we repeat the entire sequence 50 times. In summary, this Euclidean‐distance attack tests whether a held‐out PCA‐transformed data is “unusually close” to the noisy reference set compared to known non‐member points, and by averaging over multiple random trials we obtain a stable estimate of the adversary’s membership‐inference capability.
    
    
    
    \begin{figure}[htbp]
      \centering
      \begin{subfigure}[b]{0.23\textwidth}
        \centering
        \includegraphics[width=\textwidth]{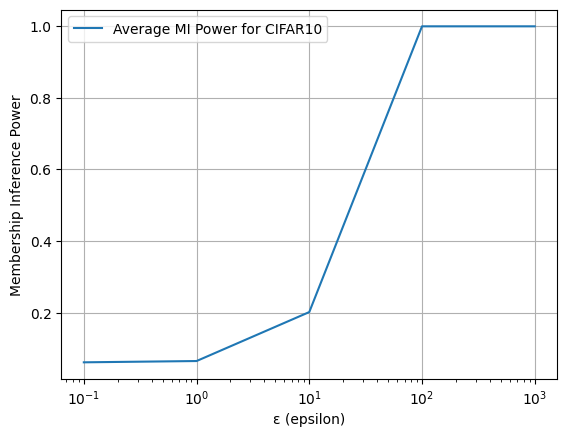}
        \caption{CIFAR-10}
        \label{fig:privacy_dataset1}
      \end{subfigure}
      \hfill
      \begin{subfigure}[b]{0.23\textwidth}
        \centering
        \includegraphics[width=\textwidth]{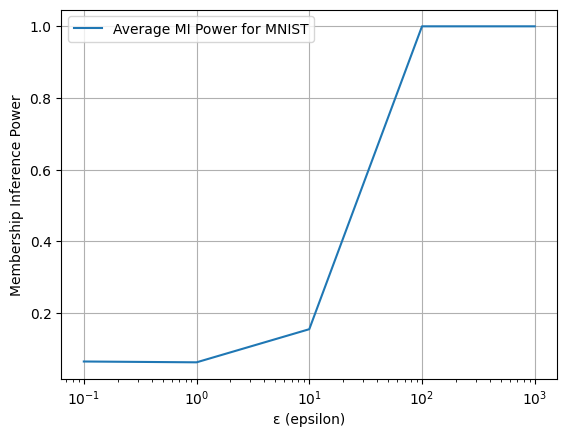}
        \caption{MNIST}
        \label{fig:privacy_dataset2}
      \end{subfigure}
      \caption{Privacy analysis results for the CIFAR-10 and MNIST datasets.}
      \label{fig:privacy_analysis}
    \end{figure}

    As shown in Figure~\ref{fig:privacy_analysis}, membership inference power is significantly influenced by the privacy parameter $\epsilon$. For both CIFAR-10 and MNIST datasets, the attack power remains consistently low (around 0.07 to 0.08) when $\epsilon$ $\leq$ 1, indicating strong privacy protection. 
    At $\epsilon = 10$, the inference power increases moderately to approximately 0.2. 
    However, a sharp escalation is observed when $\epsilon$ reaches 100 and beyond, where the attack power surges to 1.0, indicating near-perfect adversarial inference. 

    Let \(H_{0}\) denote the null hypothesis that a queried PCA vector \(x\) is not contained in the private reference set \(A\), and let \(H_{1}\) be the alternative that \(x\in A\). When the privacy budget \(\epsilon\) is large, the coordinate-wise Laplace noise added to each vector has variance orders of magnitude smaller than the natural spread of the data; consequently every released point \(\tilde a\) is almost co-located with its original \(a\in A\). Vectors drawn from outside \(A\) (non-members) remain separated from \(\widetilde{A}\) by the dataset’s intrinsic radius, so the distance distributions under \(H_{0}\) and \(H_{1}\) become nearly disjoint.
    The resulting Neyman–Pearson test therefore attains empirical power \(\approx 1\) while maintaining the false-positive rate at the fixed level \(\alpha = 0.05\). Such near-perfect separability does not arise in a black-box
    model-output attacks—widely studied in the literature—because those adversaries never observe the data itself and typically operate in a much stricter privacy regime (\(\epsilon \le 10\)).

    Based on these trends, we consider selecting $\epsilon$ values in the range of [0.1, 10], especially 10, because the EMD curve has the minimal impact by the DP noise addition when $\epsilon=10$ (see in Figure~\ref{fig:emd_results}. Also this $\epsilon$ range ensures that membership inference power remains below 0.4, thus maintaining an acceptable privacy level. 
    
    \subsection{End-to-End PQFed Performance}
        \label{sec:ete}
        
\begin{figure}[htbp]
      \centering
      \begin{subfigure}[b]{0.45\textwidth}
        \centering
        \includegraphics[width=\textwidth]{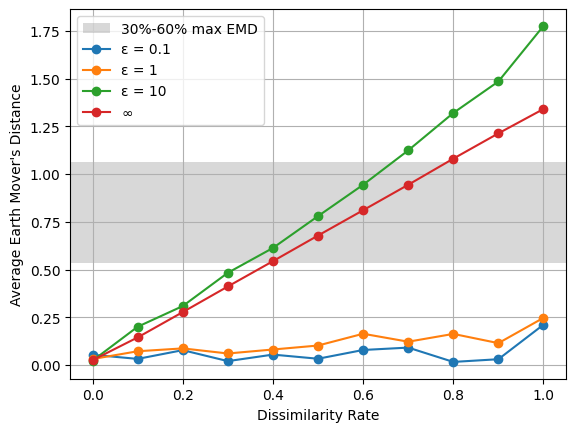}
        \caption{CIFAR-10}
        \label{fig:emd_threshold_cifar}
      \end{subfigure}
      \hfill
      \begin{subfigure}[b]{0.45\textwidth}
        \centering
        \includegraphics[width=\textwidth]{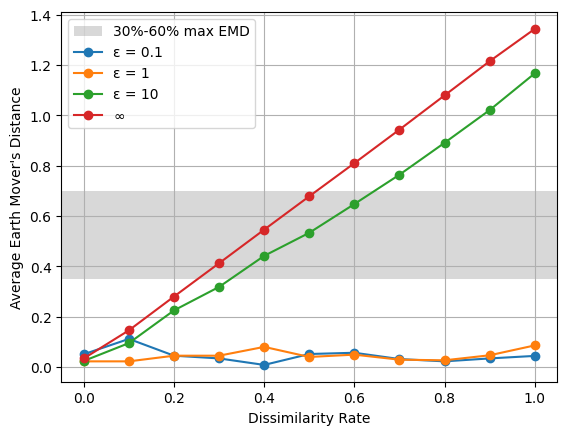}
        \caption{MNIST}
        \label{fig:emd_threshold_mnist}
      \end{subfigure}
      \caption{EMD threshold bands (30–60\% max) for (a) CIFAR-10 and (b) MNIST.}
      \label{fig:emd_thresholds}
    \end{figure}

    In real-world scenarios, clients are less likely to be strictly identical or completely dissimilar, as we have done in Section~\ref{sec:1to5}.
    Supported by the Earth Mover’s Distance (EMD) and the privacy analysis presented in Section~\ref{sec:privacy_analysis}, we simulate a real-world scenario and evaluate the end-to-end performance of our proposed scheme in identifying similar clients.
    We first construct datasets following the same procedure described in Section~\ref{sec:dissim-rate}. 
    After creating the target client $T$ as outlined in Section~\ref{sec:dataset}, we construct additional client datasets with controlled levels of dissimilarity. 
    Specifically, we introduce dissimilarity rates ranging from 0\% to 100\% in increments of 10\%, resulting in 11 client datasets.
    We then compute the EMD between each client and the target client $T$.  
    The green line in Figure~\ref{fig:emd_thresholds} shows the EMD values as the dissimilarity rate increases, with each client applying LDP with $\epsilon = 10$, as described in Section~\ref{sec:privacy_analysis}.
    Then, we manually select thresholds ranging from 30\% to 60\% of the maximum observed EMD value.  
    A strict threshold includes only clients with EMD below 30\%, meaning they are highly similar to the target client.  
    A lenient threshold allows clients with EMD up to 60\%, enabling participation from moderately similar clients.  
    Clients with EMD above 60\% are considered too dissimilar and are excluded from collaboration.  
    As a result, for both the CIFAR-10 and MNIST datasets, the target client \( T \) collaborates only with clients whose dissimilarity level falls within the 30\% threshold in the strict case.  
    In the lenient case, clients with dissimilarity levels up to 60\% are included.
    Finally, we conduct a series of FL training experiments by gradually adding clients with increasing dissimilarity rates. 
    This allows us to evaluate whether the post-noise EMD values can indicate which clients are most suitable for collaboration with client $T$.

    The experimental setup is summarized in Table~\ref{tab:fl_exp_settings}. 
    In \textbf{Experiment 6}, we use 2 clients: the target client \( T \) and one client with the same distribution. 
    In \textbf{Experiment 7}, we add one more client with 10\% dissimilarity. 
    This pattern continues incrementally: in each subsequent experiment, one additional client is added with an increasing dissimilarity level (e.g., 20\%, 30\%, ..., up to 100\%). 
    In \textbf{Experiment 16}, the setup includes 12 clients in total: the target client \( T \), one client with the same distribution, and 10 clients with dissimilarity levels ranging from 10\% to 100\% in 10\% increments.
    The results are shown in Table~\ref{tab:cifar_results} and Table~\ref{tab:mnist_results}.

    \begin{table*}[ht]
        \renewcommand{\arraystretch}{1.2}
        \centering
        \caption{Experiment Settings for End-to-End PQFed Framework.}
        \begin{tabular}{clc}
        \hline
        \textbf{Exp. Index} & \textbf{Training Set} & \textbf{Number of Clients} \\
        \hline
        6  & T + 0\% dissimilarity & 2 \\
        7  & T + 0\% dissimilarity+ 10\% dissimilarity & 3 \\
        8  & T + 0\% dissimilarity+ 10\% dissimilarity+ 20\% dissimilarity & 4 \\
        \vdots & \vdots & \vdots \\
        16 & T + 0\% + 10\% + 20\% + \dots + 100\% dissimilarity & 12 \\
        \hline
        \end{tabular}
        \label{tab:fl_exp_settings}
    \end{table*}

    \begin{table}[ht]
    \renewcommand{\arraystretch}{1.2}
    \centering
        \caption{FL Accuracy under Increasing Dissimilarity Rates (r) on CIFAR-10}
        \begin{tabular}{ccccc}
            \hline
            \textbf{Exp.} & \textbf{r (\%)} & \textbf{FedAvg (\%)} & \textbf{FedProx (\%)} & \textbf{FedDyn(\%)} \\
            \hline
            6  & 0                & 56.00 & 57.00 & 52.67 \\
            7  & 0--10            & 59.67 & 59.67 & 52.67 \\
            8  & 0--20            & 61.00 & 60.00 & 54.33 \\
            \underline{9}  & 0--30            & 60.67 & 61.33 & 56.00 \\
            10 & 0--40            & \textbf{62.33} & 62.33 & 56.00 \\
            11 & 0--50            & 61.00 & 62.33 & \textbf{56.67} \\
            \underline{12} & 0--60            & 61.00 & \textbf{64.00} & 56.33 \\
            13 & 0--70            & 61.00 & 61.67 & 54.33 \\
            14 & 0--80            & 59.67 & 60.33 & 55.67 \\
            15 & 0--90            & 59.00 & 60.33 & 53.33 \\
            16 & 0--100           & 59.67 & 59.67 & 54.33 \\
            \hline
        \end{tabular}
        \label{tab:cifar_results}
    \end{table}
    
    \begin{table}[ht]
    \renewcommand{\arraystretch}{1.2}
    \centering
    \caption{FL Accuracy under Increasing Dissimilarity Rates (r) on MNIST}
    \begin{tabular}{ccccc}
        \hline
        \textbf{Exp.} & \textbf{r (\%)} & \textbf{FedAvg (\%)} & \textbf{FedProx (\%)} & \textbf{FedDyn (\%)} \\
        \hline
        6  & 0          & 97.19 & 96.73 & 96.10 \\
        7  & 0--10       & 97.73 & 97.83 & 96.15 \\
        8  & 0--20       & 98.08 & 97.94 & 96.46 \\
        \underline{9}  & 0--30       & 98.35 & 98.40 & 96.71 \\
        10 & 0--40       & 98.56 & 98.23 & 96.33 \\
        11 & 0--50       & 98.40 & 98.44 & 96.79 \\
        \underline{12} & 0--60       & \textbf{98.71} & \textbf{98.69} & \textbf{97.00} \\
        13 & 0--70       & 98.60 & 98.67 & 96.83 \\
        14 & 0--80       & 98.46 & 98.52 & 96.94 \\
        15 & 0--90       & 98.40 & 98.67 & 96.79 \\
        16 & 0--100      & 98.44 & 98.42 & 96.42 \\
        \hline
    \end{tabular}
    \label{tab:mnist_results}
    \end{table}

    For the CIFAR-10 dataset, from Experiment 6 to 16, we observe a consistent trend across all three FL algorithms--FedAvg, FedProx, and FedDynamic--where model accuracy improves as more clients are added, then begins to degrade as too many dissimilar clients are included, with the best performance reached in Experiment 10 for FedAvg, Experiment 12 for FedProx, and Experiment 11 for FedDyn.  
    In the strict selection case, the target client \( T \) collaborates with clients up to the 30\% dissimilarity level, corresponding to Experiment 9. 
    The resulting accuracies are 60.67\%, 61.33\%, and 56.00\% for FedAvg, FedProx, and FedDyn, respectively.  
    In the lenient case, clients up to the 60\% dissimilarity level are included, corresponding to Experiment 12.  
    The resulting accuracies in this case are 61.00\%, 64.00\%, and 56.33\% for FedAvg, FedProx, and FedDyn, respectively.
    While these configurations do not always yield the absolute highest accuracy, they consistently outperform the naive aggregation baseline shown in Experiment 16, where all clients are equally included in FL.
    Considering the impact of LDP, if we examine the red line in Figure~\ref{fig:emd_thresholds}, which represents the case without applying LDP,
    the strict selection case still involves clients with up to a 30\% dissimilarity ratio.  
    However, in the lenient case, clients with up to a 70\% dissimilarity ratio are included, as shown in Experiment 13.  
    In this setting, the accuracies are 61.00\%, 61.67\%, and 54.33\% for FedAvg, FedProx, and FedDyn, respectively.  
    These results are also higher than the baseline performance reported in Experiment 16.
        
    For the MNIST dataset, similar to CIFAR-10, we observe that all three federated algorithms reach peak accuracy at a certain point, after which performance declines as more dissimilar clients are added (from Experiment 6 to 16).
    Under a strict selection, the target client $T$ collaborates only with clients up to the 30\% dissimilarity level, corresponding to Experiment 9. 
    In this setting, although FedAvg and FedProx do not surpass the baseline performance in Experiment 16, they achieve comparable results (98.35\% vs. 98.56\% for FedAvg, and 98.40\% vs. 98.56\% for FedProx).
    In the lenient case, where clients up to 60\% dissimilarity are included (Experiment 12), all 3 FL methods achieve their highest accuracy, each outperforming the baseline scenario in Experiment 16. 
    Considering the impact of LDP, we refer to the red line in Figure~\ref{fig:emd_thresholds}, which represents the non-private setting.  
    In the strict selection case, only clients with up to a 20\% dissimilarity ratio are included, as shown in Experiment 8.  
    The resulting accuracies are 98.08\%, 97.94\%, and 96.46\% for FedAvg, FedProx, and FedDyn. 
    Notably, FedAvg and FedProx also do not surpass the baseline in this setting. 
    This is similar to the performance observed along the green line. 
    In contrast, in the lenient case, clients with up to a 50\% dissimilarity ratio are included, as shown in Experiment 11.  
    The corresponding accuracies are 98.40\%, 98.44\%, and 96.79\% for FedAvg, FedProx, and FedDyn, respectively.  
    FedAvg does not exceed the baseline performance (98.40\% vs. 98.44\%), but its performance is very close.
  
    These results show that in most cases, whether under strict or lenient client selection, our pipeline successfully identifies a subset of clients that enables better global model performance compared to simply including all clients. Even when it does not surpass the performance of models trained with all clients, the results are often very close.

    \subsection{Comparison with Clustering-Based Federated Algorithm}

    As described in Section \ref{sec:framework}, PQFed utilizes a clustering algorithm to compute similarity scores between the target client and other clients, which guides the similarity-aware aggregation process. 
    This design is conceptually related to clustering-based federated personalization methods. 
    To better evaluate the effectiveness of PQFed compared to such methods, we conduct a direct comparison with a well-know clustering-based FL algorithm, IFCA \cite{ghosh2020efficient} under the same non-IID data setup.
    IFCA partitions clients into clusters and trains separate models for each cluster. 
    Specifically, each client selects a cluster model to train by minimizing its local loss, and the server maintains $k$ global models, one for each cluster. 
    In this experiment, we set the number of clusters to $k=4$, matching the K-Means clustering used in our non-IID data partitioning strategy. 
    The training procedure and hyperparameters for this experiment follow the configuration described in Section~\ref{sec:ete} (Experiment 16), with the full hyperparameter details shown in Table~\ref{tab:hyperparams}.
    \begin{table}[ht]
    \centering
    \caption{Accuracy comparison between PQFed and IFCA on CIFAR-10 and MNIST.}
    \begin{tabular}{lcc}
    \hline
    Method & Accuracy (CIFAR-10) & Accuracy (MNIST) \\
    \hline
    PQFed & 60.67--62.33\% & 98.35--98.71\% \\
    IFCA  & 61.67\%         & 97.54\%         \\
    \hline
    \end{tabular}
    \label{tab:ifca_comparison}
    \end{table}
    Table~\ref{tab:ifca_comparison} summarizes the performance comparison between PQFed and IFCA on the CIFAR-10 and MNIST datasets. 
    On MNIST, PQFed achieves an accuracy ranging from 98.35\% to 98.71\% when using different client selection thresholds (from strict to lenient), consistently outperforming IFCA, which attains 97.54\%. 
    On CIFAR-10, PQFed achieves 60.67\% to 62.33\% accuracy, while IFCA reaches 61.67\%. 
    Although IFCA slightly outperforms PQFed in some CIFAR-10 experiments, the difference is marginal, especially considering the other advantages of PQFed over IFCA that are discussed in the following.
    In addition to performance, we compare the training workflows of PQFed and IFCA. 
    In our experiments, IFCA requires all 12 clients to participate in each training round, whereas PQFed involves only a subset of clients (typically 5 to 8) selected based on their similarity to the target client, depending on whether strict or lenient selection is used.
    On the client side, IFCA requires each client to evaluate 4 local models (one for each cluster), which increases both computation and communication costs. 
    In contrast, PQFed only requires clients to train one single local model per round.
    On the server side, IFCA performs 4 separate aggregations to maintain a global model for each cluster, whereas PQFed performs just one aggregation. 
    Therefore, the proposed PQFed performs significantly more efficiently compared to IFCA.
    Additionally, as we will discuss in Section~\ref{sec:limit}, PQFed has the potential to further optimize the persobalized model of the target client by identifying and using only a subset of datasets of the other clients that align with the data distribution of the target client; whereas IFCA is limited to using the entire dataset from each client. 
    Finally, in terms of privacy, we experimentally show the robustness of PQFed against membership inference attacks in Section~\ref{sec:privacy_analysis}. 
    However, robustness of IFCA against such attacks questionable. 
    A potentially malicious server may provide very close models to the target clients in order to learn more about the training datasets of such clients by using their responses at each iteration. 

\section{Discussion}
\label{sec:dis}

    \subsection{Training Strategy Without Thresholding}
      
    Currently, the threshold is selected manually.  
    We also propose an incremental federated learning training strategy, assuming the distances between the target client \( T \) and each client are known.  
    This approach, presented in Algorithm~\ref{alg:IFL}, does not require selecting a dissimilarity threshold.
    Starting from the target client \( T \) with initial model \(\mathcal{M}_0\), we perform FL training sequentially with other clients sorted by their distance to \( T \), from closest to farthest. 
    Specifically, \(\mathcal{M}_0\) is first trained with the closest client \( c_1 \) to produce \(\mathcal{M}_1\). 
    Then, \(\mathcal{M}_1\) is used as the initial model for federated training between client \(T\) and the next closest client \(c_2\), resulting in \(\mathcal{M}_2\), and this process continues sequentially.
    After each round of training, the model is evaluated on a test set of the client \( T \).
    The training stops if the accuracy decreases compared to the previous round.

    \begin{algorithm}[htbp]
        \caption{Incremental Federated Training by Distance}
        \label{alg:IFL}
        \KwIn{
          Initial model $\mathcal{M}_0$,\\
          Ordered clients' dataset by EMD distance $\mathcal{O} = [c_1, c_2, \ldots, c_K]$,\\
          Client $T$'s train dataset $\mathcal{D}_{train}$,\\
          Client $T$' test dataset $\mathcal{D}_{test}$
        }
        \KwOut{Final model $\mathcal{M}$}
        
        $\mathcal{M} \leftarrow \mathcal{M}_0$\;
        $bestAcc \leftarrow 0$\;
        
        \For{$i \leftarrow 1$ \KwTo $K$}{
            $\mathcal{M} \leftarrow$ FederatedTrain($\mathcal{M}, \mathcal{D}_{train}, c_i$)\;
            $acc \leftarrow$ Evaluate($\mathcal{M}, \mathcal{D}_{test}$)\;
            
            \If{$acc < bestAcc$}{
                \textbf{break}\;
            }
            
            $bestAcc \leftarrow acc$\;
        }
    \Return $\mathcal{M}$\;
    \end{algorithm}

    \begin{table}[h!]
        \centering
        \caption{Incremental FL Accuracies on CIFAR-10}
        \begin{tabular}{c|c|ccc}
            \toprule
            \textbf{Round} & \textbf{r (\%)} & \textbf{FedAvg} & \textbf{FedProx} & \textbf{FedDyn} \\
            \midrule
            1  & 0   & 0.5400 & 0.5500 & 0.4800 \\
            2  & 10  & 0.6200 & 0.6233 & 0.5767 \\
            3  & 20  & 0.6567 & 0.6733 & 0.6033 \\
            4  & 30  & 0.6800 & 0.6767 & \textbf{0.6033} \\
            5  & 40  & 0.6867 & 0.7100 & 0.5867 \\
            6  & 50  & \textbf{0.7033} & 0.7167 & 0.5900 \\
            7  & 60  & 0.6967 & \textbf{0.7200} & 0.5900 \\
            8  & 70  & 0.7000 & 0.6900 & 0.5567 \\
            9  & 80  & 0.6833 & 0.7167 & 0.5733 \\
            10 & 90  & 0.6633 & 0.7000 & 0.5700 \\
            11 & 100 & 0.6667 & 0.6867 & 0.5700 \\
            \bottomrule
        \end{tabular}
        \label{tab:loop-cifar}
    \end{table}

    \begin{table}[h!]
        \centering
        \caption{Incremental FL Accuracies on MNIST}
        \begin{tabular}{c|c|ccc}
            \toprule
            \textbf{Round} & \textbf{r (\%)} & \textbf{FedAvg} & \textbf{FedProx} & \textbf{FedDyn} \\
            \midrule
            1  & 0   & 0.9685 & 0.9696 & \textbf{0.9656} \\
            2  & 10  & 0.9750 & 0.9783 & 0.9523 \\
            3  & 20  & 0.9788 & 0.9788 & 0.9456 \\
            4  & 30  & 0.9802 & 0.9815 & 0.9308 \\
            5  & 40  & 0.9808 & 0.9840 & 0.9285 \\
            6  & 50  & 0.9808 & 0.9825 & 0.9263 \\
            7  & 60  & 0.9827 & \textbf{0.9846} & 0.9119 \\
            8  & 70  & \textbf{0.9831} & 0.9840 & 0.9190 \\
            9  & 80  & 0.9831 & 0.9846 & 0.9208 \\
            10 & 90  & 0.9833 & 0.9840 & 0.9123 \\
            11 & 100 & 0.9821 & 0.9831 & 0.9142 \\
            \bottomrule
        \end{tabular}
        \label{tab:loop-mnist}
    \end{table}

    We also propose an incremental federated learning training strategy, assuming the distances between the target client \( T \) and each client are known.  
    This approach, presented in Algorithm~\ref{alg:IFL}, does not require selecting a dissimilarity threshold.  
    We apply our proposed algorithm to the experiments described in Section~\ref{sec:ete}, and report the results in Table~\ref{tab:loop-cifar} and Table~\ref{tab:loop-mnist}.
    
    For CIFAR-10, the algorithm stops at round 4 for FedAvg, round 7 for FedProx, and round 6 for FedDyn.  
    In all cases, it achieves higher performance compared to the baseline setup in Experiment 16 \ref{sec:ete}.  
    Moveover, compared to the highest accuracies previously observed (62.33\%, 64.00\%, and 56.67\% for FedAvg, FedProx, and FedDyn, respectively), our algorithm achieves significantly better results, surpassing each of them.
    For MNIST, the algorithm converges at rounds 7, 8, and 1 for FedProx, FedDyn, and FedAvg, respectively.  
    While not all methods achieve the absolute highest accuracy, both FedProx and FedDyn outperform the baseline in Experiment 16, and FedAvg achieves a performance that is very close (98.31\% vs. 98.56\%).
    In addition, under this strategy, PQFed outperforms IFCA on both datasets, achieving 70.33\% vs. 61.67\% on CIFAR-10 and 98.31\% vs. 97.54\% on MNIST.

    In future work, a more principled analysis of the trade-offs between performance and communication cost of this algorithm could be explored.

    \subsection{ Robustness to Dissimilar Datasets} 
    \label{sec:balance}
    
    We assume that introducing more dissimilar datasets in FL can lead to degraded performance for the target client $T$. 
    However, in some scenarios, this effect is not clearly observed.
    As shown in Experiments 1 to 5 on the MNIST dataset (see Table ~\ref{tab:1t5results}), although the results generally support our hypothesis, the performance differences are relatively small.
    We attribute this to the simplicity of the MNIST dataset and the effectiveness of CNNs on it. 
    Even with limited clean data, CNN models are able to learn effectively and achieve strong results \cite{kadam2020cnn}.
    For such simple datasets, the global model appears to be tolerant of a degree of data diversity. 
    As a result, a larger number of clients can be included without significantly affecting performance. 
    This finding is further supported by Experiments 6 to 16 (see Table \ref{tab:cifar_results}).
    While Experiment 12 achieves the highest performance, all results from Experiments 7 to 16 remain closely aligned, indicating dissimilarity data do not significantly impact the global model's accuracy.
    Even with a strict selection threshold, as in Experiment 9, the performance does not significantly surpass that of including all clients, as in Experiment 16. 
    Nevertheless, PQFed effectively reduces training costs by limiting participation to only the most relevant clients—for example, reducing the number of active clients from 12 to 5.
         
    \subsection{Impact of Differential Privacy Noise on EMD Trends}
    \label{sec:dp_on_emd}
    The distinct trends observed in the EMD curves, shown in Figure~\ref{fig:emd_results}, at different privacy settings can be explained through the nature of DP noise addition in combination with PCA-based feature transformations.

    Under strong differential privacy settings (low $\epsilon$ values, e.g., 0.1 or 1), the magnitude of the noise added is significantly larger due to the stringent privacy requirements. This substantial noise causes PCA-transformed data points to shift drastically from their original locations in the reduced-dimensional space. Because PCA compresses the majority of data variance into a few principal components, even modest perturbations in the original high-dimensional space, amplified by low-$\epsilon$ DP mechanisms, translate into large and unpredictable displacements in the PCA feature space. As a consequence, data points become scattered more uniformly throughout this space, diminishing the inherent cluster structure and creating highly overlapping distributions among clients. Such uniformly dispersed points result in distributions across different clients becoming similarly "chaotic," and therefore more indistinguishable. This phenomenon directly explains why EMD curves remain relatively flat under strong DP protection: the added noise overwhelms any genuine distributional shifts that would otherwise emerge from increasing out-of-cluster samples, causing all datasets to appear equally mixed or randomized.

    In contrast, under weaker differential privacy settings (higher $\epsilon$, e.g., 10) or the non-private scenario ($\epsilon = \infty$), the magnitude of noise is comparatively smaller or absent, allowing the PCA-transformed data points to retain clearer traces of their original structural information. Here, differences in data distribution—particularly introduced by higher proportions of out-of-cluster samples—are preserved and clearly reflected in the EMD measurements. This preservation of structural fidelity ensures that EMD accurately detects incremental distributional changes, producing a visible upward trend as the dissimilarity rate grows. Such moderate privacy settings balance the competing needs of protecting individual privacy and maintaining meaningful data structure.

    Thus, the flattening of EMD trends observed at lower epsilon values is not merely a statistical artifact but an expected consequence of the extensive data perturbation applied for strong privacy guarantees. This reasoning supports careful selection of privacy levels that adequately balance meaningful data distribution assessments with effective privacy protections in FL settings.

    \subsection{Clustering-Based Dataset Similarity}
    \label{sec:limit}
    Our pipeline performs final evaluations based on distance similarity, which is influenced by the clustering model trained on the central server $S$. 
    We use clustering distribution, the proportion of a client’s data assigned to each cluster after K-means—as a proxy for dataset similarity (For example, if 30\% of a client's data belongs to cluster 2, 20\% to cluster 5, and 50\% to cluster 12, this defines its clustering distribution).
    When 2 datasets exhibit similar clustering distributions, their label distributions, the proportion of samples per class label, also tend to align. 
    This is because each cluster typically captures a consistent mix of labels, grouping semantically similar data points. 
    Our experiments confirm that matching clustering distributions lead to more consistent label distributions, contributing to improved and more stable global model performance.
    These findings align with prior work \cite{DBLP:journals/corr/abs-2102-02079}, which emphasizes that similarity in label distributions across clients facilitates better convergence of the global model. 
    In our approach, clustering-based distance metrics implicitly capture this alignment, enhancing evaluation quality.
    However, in real-world scenarios, it is often difficult to find clients whose cluster distributions exactly match that of the target client. 
    One possible extension is to select subsets from each client’s dataset to better align with the cluster distribution of the target client. 
    Since it reduces the size of available training data, we plan to explore in future work how this trade-off affects model performance.

    \subsection{Clarification on Client-Specific Membership Inference Risk vs. Average Risk}
\begin{figure}[htbp]
    \centering
    \begin{subfigure}[b]{0.23\textwidth}
        \centering
        \includegraphics[width=\textwidth]{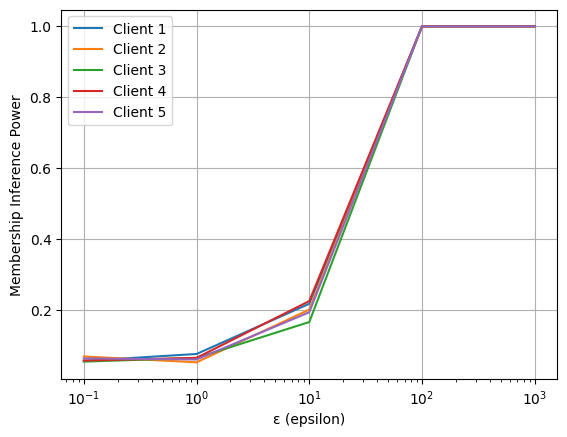}
        \caption{CIFAR-10: Client-specific membership inference power}
        \label{fig:privacy_analysis_cifar10_5lines}
    \end{subfigure}
    \hfill
    \begin{subfigure}[b]{0.23\textwidth}
        \centering
        \includegraphics[width=\textwidth]{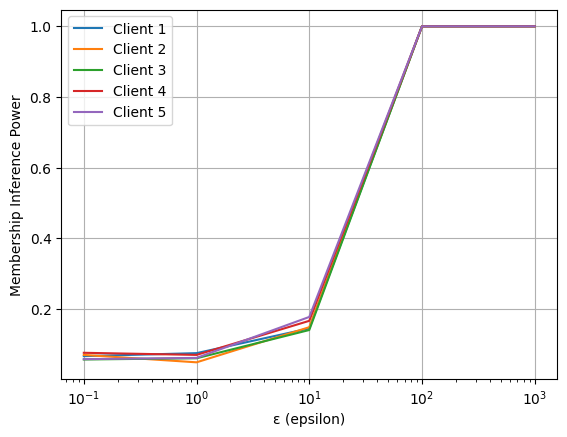}
        \caption{MNIST: Client-specific membership inference power}
        \label{fig:privacy_analysis_mnist_5lines}
    \end{subfigure}
    \caption{Membership inference power across individual clients under different $\epsilon$ values.}
    \label{fig:privacy_analysis_5lines}
    \end{figure}

    In the privacy analysis section ~\ref{sec:privacy_analysis}, we previously reported membership inference attack power using averaged results across all participating clients. 
    Here, in Figure~\ref{fig:privacy_analysis_cifar10_5lines} (CIFAR-10) and Figure~\ref{fig:privacy_analysis_mnist_5lines} (MNIST), we provide an additional analysis depicting membership inference risks separately for each client to validate our earlier averaging approach.
    As illustrated by these detailed client-specific analyses, the membership inference power across individual clients remains consistently close, with minimal deviation, especially at stronger privacy settings ($\epsilon \leq 10$). 
    This observation supports the validity and appropriateness of using the average membership inference power as a representative measure in the earlier privacy analysis. 
    The close alignment among individual client curves indicates that no single client disproportionately influences the overall privacy risk estimation, reinforcing our conclusion that the averaged results effectively capture the general trend and magnitude of privacy risks under various differential privacy guarantees.

    Thus, presenting averaged membership inference power in our initial analyses remains justified and methodologically sound, as the underlying variation among clients is minor. This supplementary client-specific analysis serves primarily as additional validation and clarification to preemptively address potential concerns regarding the representativeness of our reported averaged results.

\section{Conclusion}
We present PQFed, a personalized federated learning framework that incorporates an early-stage, privacy-preserving quality control mechanism to enable client collaboration. 
PQFed performs feature extraction and clustering, using Earth Mover’s Distance to estimate dataset similarity across clients and set thresholds for selecting potential collaborators. 
It employs local differential privacy to defend against membership inference attacks, ensuring that sensitive client data remains protected throughout the process.
We implement PQFed on 3 classic federated learning algorithms, including FedAvg, FedProx, and FedDyn, and evaluate it on MNIST and CIFAR-10. 
Additionally, we compare PQFed with IFCA algorithm, to highlight its advantages over traditional cluster-based federated learning methods. 
The results demonstrate that PQFed provides strong privacy guarantees while maintaining high personalization performance. 
By identifying and excluding low-quality or incompatible datasets prior to training, PQFed also reduces communication and computation costs.
Overall, PQFed is a secure and effective personalized approach that enhances efficiency in addressing data heterogeneity challenges in federated learning.

\bibliographystyle{IEEEtran}
\bibliography{IEEEtran}

\begin{thebibliography}{10}
\providecommand{\url}[1]{#1}
\csname url@samestyle\endcsname
\providecommand{\newblock}{\relax}
\providecommand{\bibinfo}[2]{#2}
\providecommand{\BIBentrySTDinterwordspacing}{\spaceskip=0pt\relax}
\providecommand{\BIBentryALTinterwordstretchfactor}{4}
\providecommand{\BIBentryALTinterwordspacing}{\spaceskip=\fontdimen2\font plus
\BIBentryALTinterwordstretchfactor\fontdimen3\font minus \fontdimen4\font\relax}
\providecommand{\BIBforeignlanguage}[2]{{%
\expandafter\ifx\csname l@#1\endcsname\relax
\typeout{** WARNING: IEEEtran.bst: No hyphenation pattern has been}%
\typeout{** loaded for the language `#1'. Using the pattern for}%
\typeout{** the default language instead.}%
\else
\language=\csname l@#1\endcsname
\fi
#2}}
\providecommand{\BIBdecl}{\relax}
\BIBdecl

\bibitem{mcmahan2017communication}
B.~McMahan, E.~Moore, D.~Ramage, S.~Hampson, and B.~A. y~Arcas, ``Communication-efficient learning of deep networks from decentralized data,'' in \emph{Artificial intelligence and statistics}.\hskip 1em plus 0.5em minus 0.4em\relax PMLR, 2017, pp. 1273--1282.

\bibitem{DBLP:journals/corr/abs-2102-02079}
\BIBentryALTinterwordspacing
Q.~Li, Y.~Diao, Q.~Chen, and B.~He, ``Federated learning on non-iid data silos: An experimental study,'' \emph{CoRR}, vol. abs/2102.02079, 2021. [Online]. Available: \url{https://arxiv.org/abs/2102.02079}
\BIBentrySTDinterwordspacing

\bibitem{ye2023heterogeneous}
M.~Ye, X.~Fang, B.~Du, P.~C. Yuen, and D.~Tao, ``Heterogeneous federated learning: State-of-the-art and research challenges,'' \emph{ACM Computing Surveys}, vol.~56, no.~3, pp. 1--44, 2023.

\bibitem{li2020fedprox}
\BIBentryALTinterwordspacing
T.~Li, A.~K. Sahu, M.~Zaheer, M.~Sanjabi, A.~Talwalkar, and V.~Smith, ``Federated optimization in heterogeneous networks,'' 2020. [Online]. Available: \url{https://arxiv.org/abs/1812.06127}
\BIBentrySTDinterwordspacing

\bibitem{acar2021federatedlearningbaseddynamic}
\BIBentryALTinterwordspacing
D.~A.~E. Acar, Y.~Zhao, R.~M. Navarro, M.~Mattina, P.~N. Whatmough, and V.~Saligrama, ``Federated learning based on dynamic regularization,'' 2021. [Online]. Available: \url{https://arxiv.org/abs/2111.04263}
\BIBentrySTDinterwordspacing

\bibitem{fallah2020personalizedfederatedlearningmetalearning}
\BIBentryALTinterwordspacing
A.~Fallah, A.~Mokhtari, and A.~Ozdaglar, ``Personalized federated learning: A meta-learning approach,'' 2020. [Online]. Available: \url{https://arxiv.org/abs/2002.07948}
\BIBentrySTDinterwordspacing

\bibitem{hu2020personalized}
R.~Hu, Y.~Guo, H.~Li, Q.~Pei, and Y.~Gong, ``Personalized federated learning with differential privacy,'' \emph{IEEE Internet of Things Journal}, vol.~7, no.~10, pp. 9530--9539, 2020.

\bibitem{LiSample-level}
A.~Li, L.~Zhang, J.~Tan, Y.~Qin, J.~Wang, and X.-Y. Li, ``Sample-level data selection for federated learning,'' in \emph{IEEE INFOCOM 2021 - IEEE Conference on Computer Communications}, 2021, pp. 1--10.

\bibitem{shi2023makelandscapeflatterdifferentially}
\BIBentryALTinterwordspacing
Y.~Shi, Y.~Liu, K.~Wei, L.~Shen, X.~Wang, and D.~Tao, ``Make landscape flatter in differentially private federated learning,'' 2023. [Online]. Available: \url{https://arxiv.org/abs/2303.11242}
\BIBentrySTDinterwordspacing

\bibitem{xuSAFE}
X.~Xu, H.~Li, Z.~Li, and X.~Zhou, ``Safe: Synergic data filtering for federated learning in cloud-edge computing,'' \emph{IEEE Transactions on Industrial Informatics}, vol.~19, no.~2, pp. 1655--1665, 2023.

\bibitem{sattler2019clusteredfederatedlearningmodelagnostic}
\BIBentryALTinterwordspacing
F.~Sattler, K.-R. Müller, and W.~Samek, ``Clustered federated learning: Model-agnostic distributed multi-task optimization under privacy constraints,'' 2019. [Online]. Available: \url{https://arxiv.org/abs/1910.01991}
\BIBentrySTDinterwordspacing

\bibitem{ghosh2020efficient}
A.~Ghosh, J.~Chung, D.~Yin, and K.~Ramchandran, ``An efficient framework for clustered federated learning,'' \emph{Advances in neural information processing systems}, vol.~33, pp. 19\,586--19\,597, 2020.

\bibitem{yue2024phase}
W.~Yue, P.~K. Tripathi, G.~Ponon, Z.~Ualikhankyzy, D.~W. Brown, B.~Clausen, M.~Strantza, D.~C. Pagan, M.~A. Willard, F.~Ernst \emph{et~al.}, ``Phase identification in synchrotron x-ray diffraction patterns of ti--6al--4v using computer vision and deep learning,'' \emph{Integrating Materials and Manufacturing Innovation}, vol.~13, no.~1, pp. 36--52, 2024.

\bibitem{Dervishi2023}
L.~Dervishi, W.~Li, A.~Halimi, X.~Jiang, J.~Vaidya, and E.~Ayday, ``\BIBforeignlanguage{en}{Privacy preserving identification of population stratification for collaborative genomic research},'' \emph{\BIBforeignlanguage{en}{Bioinformatics}}, vol.~39, no. 39 Suppl 1, pp. i168--i176, Jun. 2023.

\bibitem{shin2022fedbalancerdatapacecontrol}
\BIBentryALTinterwordspacing
J.~Shin, Y.~Li, Y.~Liu, and S.-J. Lee, ``Fedbalancer: Data and pace control for efficient federated learning on heterogeneous clients,'' 2022. [Online]. Available: \url{https://arxiv.org/abs/2201.01601}
\BIBentrySTDinterwordspacing

\bibitem{shyn2021fedccea}
S.~K. Shyn, D.~Kim, and K.~Kim, ``Fedccea : A practical approach of client contribution evaluation for federated learning,'' 2021.

\bibitem{guo2023fedbrimprovingfederatedlearning}
\BIBentryALTinterwordspacing
Y.~Guo, X.~Tang, and T.~Lin, ``Fedbr: Improving federated learning on heterogeneous data via local learning bias reduction,'' 2023. [Online]. Available: \url{https://arxiv.org/abs/2205.13462}
\BIBentrySTDinterwordspacing

\bibitem{li2023fedsdgfsefficientsecurefeature}
\BIBentryALTinterwordspacing
A.~Li, H.~Peng, L.~Zhang, J.~Huang, Q.~Guo, H.~Yu, and Y.~Liu, ``Fedsdg-fs: Efficient and secure feature selection for vertical federated learning,'' 2023. [Online]. Available: \url{https://arxiv.org/abs/2302.10417}
\BIBentrySTDinterwordspacing

\bibitem{Reputation-Aware}
X.~Tan, W.~C. Ng, W.~Y.~B. Lim, Z.~Xiong, D.~Niyato, and H.~Yu, ``Reputation-aware federated learning client selection based on stochastic integer programming,'' \emph{IEEE Transactions on Big Data}, vol.~10, no.~6, pp. 953--964, 2024.

\bibitem{li2020federated}
T.~Li, A.~K. Sahu, A.~Talwalkar, and V.~Smith, ``Federated learning: Challenges, methods, and future directions,'' \emph{IEEE signal processing magazine}, vol.~37, no.~3, pp. 50--60, 2020.

\bibitem{haddadpour2019convergencelocaldescentmethods}
\BIBentryALTinterwordspacing
F.~Haddadpour and M.~Mahdavi, ``On the convergence of local descent methods in federated learning,'' 2019. [Online]. Available: \url{https://arxiv.org/abs/1910.14425}
\BIBentrySTDinterwordspacing

\bibitem{abdi2010principal}
H.~Abdi and L.~J. Williams, ``Principal component analysis,'' \emph{Wiley interdisciplinary reviews: computational statistics}, vol.~2, no.~4, pp. 433--459, 2010.

\bibitem{sppca}
D.~Froelicher, H.~Cho, M.~Edupalli, J.~Sa~Sousa, J.-P. Bossuat, A.~Pyrgelis, J.~R. Troncoso-Pastoriza, B.~Berger, and J.-P. Hubaux, ``Scalable and privacy-preserving federated principal component analysis,'' in \emph{2023 IEEE Symposium on Security and Privacy (SP)}, 2023, pp. 1908--1925.

\bibitem{dispca}
S.~X. Wu, H.-T. Wai, L.~Li, and A.~Scaglione, ``A review of distributed algorithms for principal component analysis,'' \emph{Proceedings of the IEEE}, vol. 106, no.~8, pp. 1321--1340, 2018.

\bibitem{lloyd1982least}
S.~Lloyd, ``Least squares quantization in pcm,'' \emph{IEEE transactions on information theory}, vol.~28, no.~2, pp. 129--137, 1982.

\bibitem{rubner2000earth}
Y.~Rubner, C.~Tomasi, and L.~J. Guibas, ``The earth mover's distance as a metric for image retrieval,'' \emph{International journal of computer vision}, vol.~40, pp. 99--121, 2000.

\bibitem{NEURIPS2020_f52a7b26}
D.~Alvarez-Melis and N.~Fusi, ``Geometric dataset distances via optimal transport,'' in \emph{Advances in Neural Information Processing Systems}, H.~Larochelle, M.~Ranzato, R.~Hadsell, M.~Balcan, and H.~Lin, Eds., vol.~33.\hskip 1em plus 0.5em minus 0.4em\relax Curran Associates, Inc., 2020, pp. 21\,428--21\,439.

\bibitem{Yeom2017-ho}
S.~Yeom, I.~Giacomelli, M.~Fredrikson, and S.~Jha, ``Privacy risk in machine learning: Analyzing the connection to overfitting,'' \emph{arXiv [cs.CR]}, Sep. 2017.

\bibitem{Shokri2017-ps}
R.~Shokri, M.~Stronati, C.~Song, and V.~Shmatikov, ``Membership inference attacks against machine learning models,'' in \emph{2017 IEEE Symposium on Security and Privacy (SP)}, May 2017, pp. 3--18.

\bibitem{Carlini2022-nw}
N.~Carlini, S.~Chien, M.~Nasr, S.~Song, A.~Terzis, and F.~Tramèr, ``Membership inference attacks from first principles,'' in \emph{2022 IEEE Symposium on Security and Privacy (SP)}.\hskip 1em plus 0.5em minus 0.4em\relax IEEE, May 2022.

\bibitem{truex2020ldp}
S.~Truex, L.~Liu, K.-H. Chow, M.~E. Gursoy, and W.~Wei, ``Ldp-fed: Federated learning with local differential privacy,'' in \emph{Proceedings of the third ACM international workshop on edge systems, analytics and networking}, 2020, pp. 61--66.

\bibitem{huang2020improving}
W.~Huang, S.~Zhou, T.~Zhu, Y.~Liao, C.~Wu, and S.~Qiu, ``Improving laplace mechanism of differential privacy by personalized sampling,'' in \emph{2020 IEEE 19th international conference on trust, security and privacy in computing and communications (TrustCom)}.\hskip 1em plus 0.5em minus 0.4em\relax IEEE, 2020, pp. 623--630.

\bibitem{jiang2021federated}
Y.~Jiang, S.~Wang, V.~Valls, B.~J. Ko, W.~Lee, and K.~K. Leung, ``Federated learning with differential privacy: Algorithms and performance analysis,'' \emph{IEEE Transactions on Information Forensics and Security}, vol.~16, pp. 3454--3469, 2021.

\bibitem{Halimi2022}
A.~Halimi, L.~Dervishi, E.~Ayday, A.~Pyrgelis, J.~R. Troncoso-Pastoriza, J.-P. Hubaux, X.~Jiang, and J.~Vaidya, ``Privacy-preserving and efficient verification of the outcome in genome-wide association studies,'' \emph{arXiv [cs.CR]}, Jan. 2021.

\bibitem{krizhevsky2009learning}
A.~Krizhevsky, ``Learning multiple layers of features from tiny images,'' University of Toronto, Technical Report UTML TR 2009-001, 2009.

\bibitem{mnist}
Y.~Lecun, L.~Bottou, Y.~Bengio, and P.~Haffner, ``Gradient-based learning applied to document recognition,'' \emph{Proceedings of the IEEE}, vol.~86, no.~11, pp. 2278--2324, 1998.

\bibitem{kadam2020cnn}
S.~S. Kadam, A.~C. Adamuthe, and A.~B. Patil, ``Cnn model for image classification on mnist and fashion-mnist dataset,'' \emph{Journal of scientific research}, vol.~64, no.~2, pp. 374--384, 2020.

\end{thebibliography}

\end{document}